\documentclass[lettersize,journal, dvipsnames]{IEEEtran}
\usepackage{amsmath,amsfonts}
\usepackage{algorithmic}
\usepackage{algorithm}
\usepackage{array}
\usepackage{subfig}
\usepackage{textcomp}
\usepackage{stfloats}
\usepackage{url}
\usepackage{verbatim}
\usepackage{graphicx}
\usepackage{overpic}
\usepackage{cite}
\usepackage{hyperref}
\usepackage{authblk}
\usepackage{xcolor}
\usepackage{multirow}
\usepackage{tabularx}  
\usepackage{amssymb, amsthm}

\newcommand{\UPH}{\mathrm{UPH}}

\begin{document}

\title{Stow: Robotic Packing of Items into Fabric Pods}

\author{Nicolas~Hudson, Josh~Hooks, Rahul~Warrier, Curt~Salisbury, Ross~Hartley, Kislay~Kumar, Bhavana~Chandrashekhar, Paul~Birkmeyer, Bosch~Tang, Matt~Frost, Shantanu~Thakar, Tony~Piaskowy, Petter~Nilsson, Josh~Petersen, Neel~Doshi, Alan~Slatter, Ankit~Bhatia, Cassie~Meeker, Yuechuan~Xue, Dylan~Cox, Alex~Kyriazis, Bai~Lou, Nadeem~Hasan, Asif~Rana, Nikhil~Chacko, Ruinian~Xu, Siamak~Faal, Esi~Seraj, Mudit~Agrawal, Kevin~Jamieson, Alessio~Bisagni, Valerie~Samzun, Christine~Fuller, Alex~Keklak, Alex~Frenkel, Lillian~Ratliff, Aaron~Parness\thanks{L. Ratliff and K. Jamieson are joint with Amazon Robotics and University of Washington.}}
\affil{Amazon~Robotics}
%


\maketitle

\begin{abstract}
This paper presents a compliant manipulation system capable of placing items onto densely packed shelves. The wide diversity of items and strict business requirements for high producing rates and low defect generation have prohibited warehouse robotics from performing this task. Our innovations in hardware, perception, decision-making, motion planning, and control have enabled this system to perform over 500,000 stows in a large e-commerce fulfillment center. The system achieves human levels of packing density and speed while prioritizing work on overhead shelves to enhance the safety of humans working alongside the robots. Supplementary video provides an overview of this work.
\end{abstract}


\section{Introduction}

\begin{figure}[t!]
\centering
\includegraphics[width=1.0\linewidth]{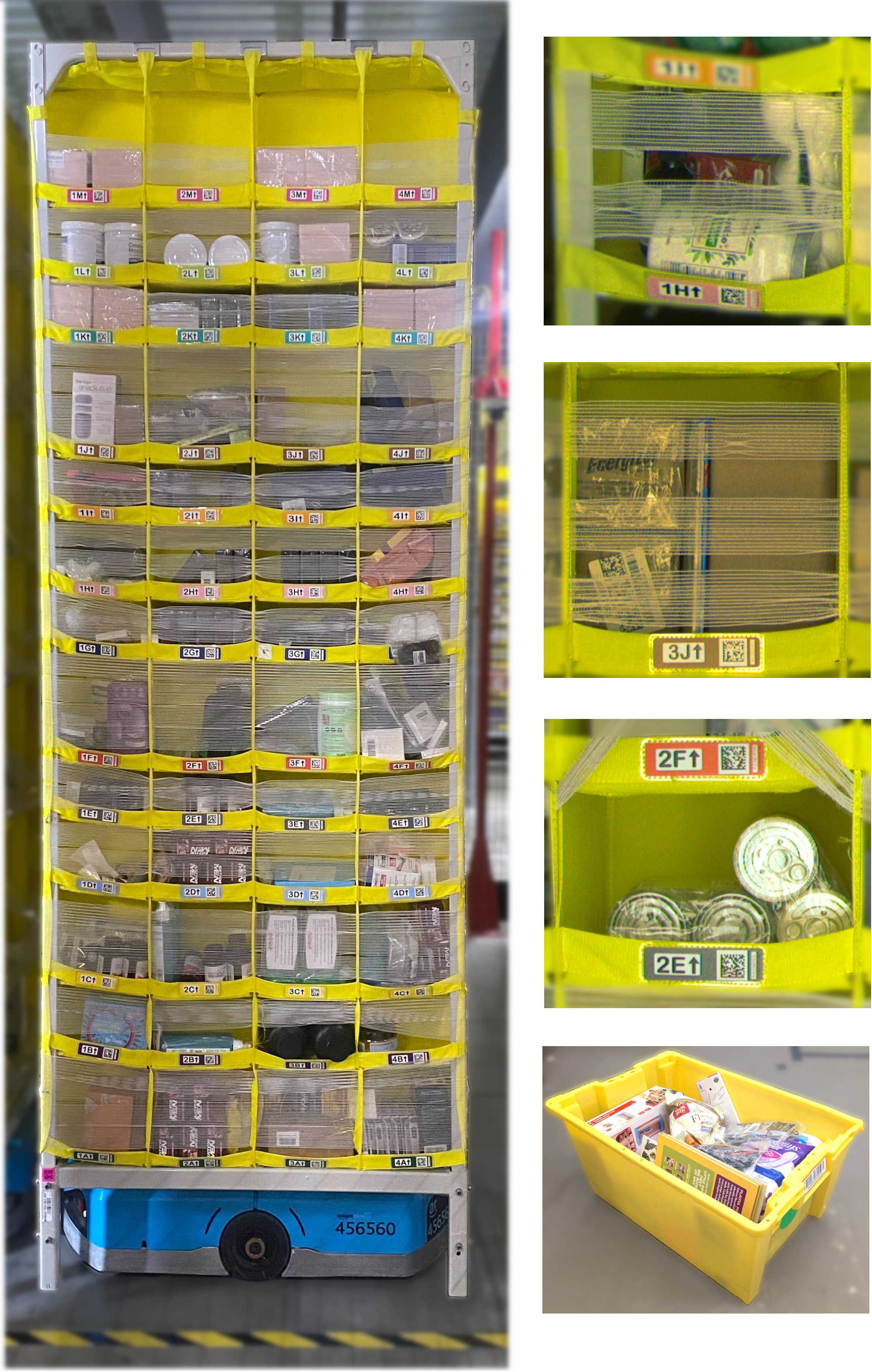}
\caption{Storage Pod: Items are stored densely and heterogeneously within each bin on this pod face. Elastic bands are used to prevent objects from falling out as pods are moved throughout the building by mobile robots. Bins are continuously picked from and stowed into during operations. Items arrive at stow stations in plastic totes, bottom right.}
\label{fig:StoragePod}
\end{figure}

\IEEEPARstart{S}{towing} is the task of placing inventory onto fabric shelves for storage. Individual items are stowed within fabric pods, four-sided yellow bookcases partitioned into bins (see Fig. \ref{fig:StoragePod}). Items are stowed heterogeneously to improve the density of storage and the speed of access to a given item for which there may be many copies in the storage field. Mobile robots operating on a grid of fiducials move pods to stations around the perimeter of the storage field for stowing and picking \cite{wurman2008coordinating}. Due to the wide range of items (more than a million unique items in a single warehouse) and the desire to store them densely, these tasks require operating in a high degree of clutter and a high degree of contact, something robots historically struggle to accomplish reliably. Further complicating the task, semi-transparent elastic retention bands span the front of the bins to prevent items from falling out as the mobile robots move the pods around the building. These elastic bands require additional manipulation steps and complicate the use of off-the-shelf 3D perception solutions. The task is currently performed manually more than 14 billion times per year.

The robotic solution described here is designed to stow 80\% of items in the warehouse at a rate of 300 units per hour. It must achieve the same high-density storage that the warehouses currently rely on to provide customer selection, and it must be capable of operating more than 20 hr per day and 7 days per week. New products are continuously introduced, so a priori information on item properties cannot be guaranteed. In addition to the elastic bands, the walls of the bin and a 1-inch tall bin lip provide collision volumes that prevent robot poses and camera views typically used in open countertop demos of new robotic manipulation systems.

\section{Related Works} 

The 2015-2017 Amazon Robotic Challenges (ARC) \cite{Corbato2018ARC,Correll2018ARC,Morrison2018Cartman} solicited exploration of this problem by asking teams to pick diverse items from a tote and stow them onto shelves or into other containers. Deployment in an e-commerce warehouse required solving several additional challenges not included in the ARC.  First, the diversity of items is far greater (millions vs tens), and many item attributes including shape, color and compliance, are not known ahead of time, and basic object information like mass and dimensions are only approximate. Second, each item must be stored in a state that makes it visually and physically accessible while maximizing storage density. Third, the stow robot must also avoid inducing \emph{damage} or causing items to fall to the floor (called \emph{amnesty}). 

These problems of placement etiquette, amnesty and damage are less studied in academic literature, but are important for production systems. Robotic object placement has been studied in the context of clearing space in clutter \cite{Cosgun2011Push} or using semantics \cite{Basu2012Place} for in-home tasks and predicting object motion and stability while stowing on a shelf \cite{Chen2023Predict}. Object stacking \cite{Li2017Stability, Hundt2020GoodRobot, Lee2021BeyondPickAndPlace}, or placing one object on top of another, has also been studied. The system presented here prioritizes vertical stowing in a bookshelf like manner and only admits stacking via simple heuristics in limited cases as stacked items can slip out between retaining bands during pod motion. Dropping objects has been well studied in the context of grasping \cite{Newbury2023}, but typically stability is evaluated during free space moves or from form or force closure metrics \cite{Lynch2017ModernRobotics}. The primary cause of amnesty here is during placement, where it either catches on the elastic (see Figure \ref{fig:StoragePod}), or item-to-item dynamics, where catching on compressing adjacent items causes items to falls out of the bin. Damaging items while manipulating them is reasonably understood for suction grippers \cite{Aduh2024ICRA}, where non-prehensile suction grasps can deconstruct boxes and fold/crease books when grasping their covers. Pinch grasping typically avoids these specific problems, and careful mechanical design is used here to avoid damage both when grasping and inserting objects.

Robot morphology plays an important role in simplifying dense packing. Most ARC teams used a combination of suction and pinch grippers to enable robust grasping and placement of diverse items \cite{Corbato2018ARC, Correll2018ARC, Morrison2018Cartman}. We have found end effector design to be critical achieving dense packing. Recent work on consolidating items has used a single arm to grasp, make space in clutter, and place \cite{Ai2024Robopack, Chen2023Predict}. This work is impressive in that it seeks to learn item-to-item interactions and robust reactive behaviors. This prediction and control problem can be simplified by using dual manipulators \cite{Manyar2024}, using one arm to with a paddle to push items and make space, and a second arm with a pinch gripper to place items. The stow robot instead combines these two manipulators into a specialized end effector with an extendable plank to push in bin items.

Bin Packing, or choosing which item and where to place it, is a long studied problem \cite{Lodi2002AppliedMath} which seeks to optimally assign items into a finite set of bins. Recent versions of this optimization problem have represented deformable objects \cite{Zuo20223DBPPDeformable} as maximum compressibilty constraints on cuboids, but this work does not consider the robotic behaviors or motion required to achieve that packing.  Robotics literature has considered packing novel items \cite{Wang2022Packing}, taking into account the feasibility of motions and using behaviors such as pushing to increase density and achieve alignment \cite{Shome2019RobustPacking}. Recent work has used diffusion models to learn where objects should be placed \cite{Liu2023structdiffusion, Yang2023diffusion} including where to place objects on shelves \cite{Simeonov2023shelvingstacking}, but these do not yet consider moving or pushing already present items nor their deformation. Graph Neural Networks have been used to predict the motion of items \cite{Ai2024Robopack, Chen2023Predict} given the robotic action. In contrast to these prior works, we find the salient problem for this tight fit bin packing problem is primarily about non-contact (i.e., perception driven) prediction of available space and prior estimation of behavioral success given the in-hand object and perceived bin state. This is primarily driven by needing to increase the number of items stored per hour, which limits the number of unproductive cycles. In these cycles the robot moves items in the bin, measures the space kinesthetically, but is unable to make enough space to stow the target item. Posing the problem as success prediction is akin to 'sampling with a learned metric' as described in a recent survey on learned grasping \cite{Newbury2023}, and here we learn this metric from real world data at scale \cite{Li2023Robin}.

\section{Robotic Stow Process}\label{sec:stow-process}

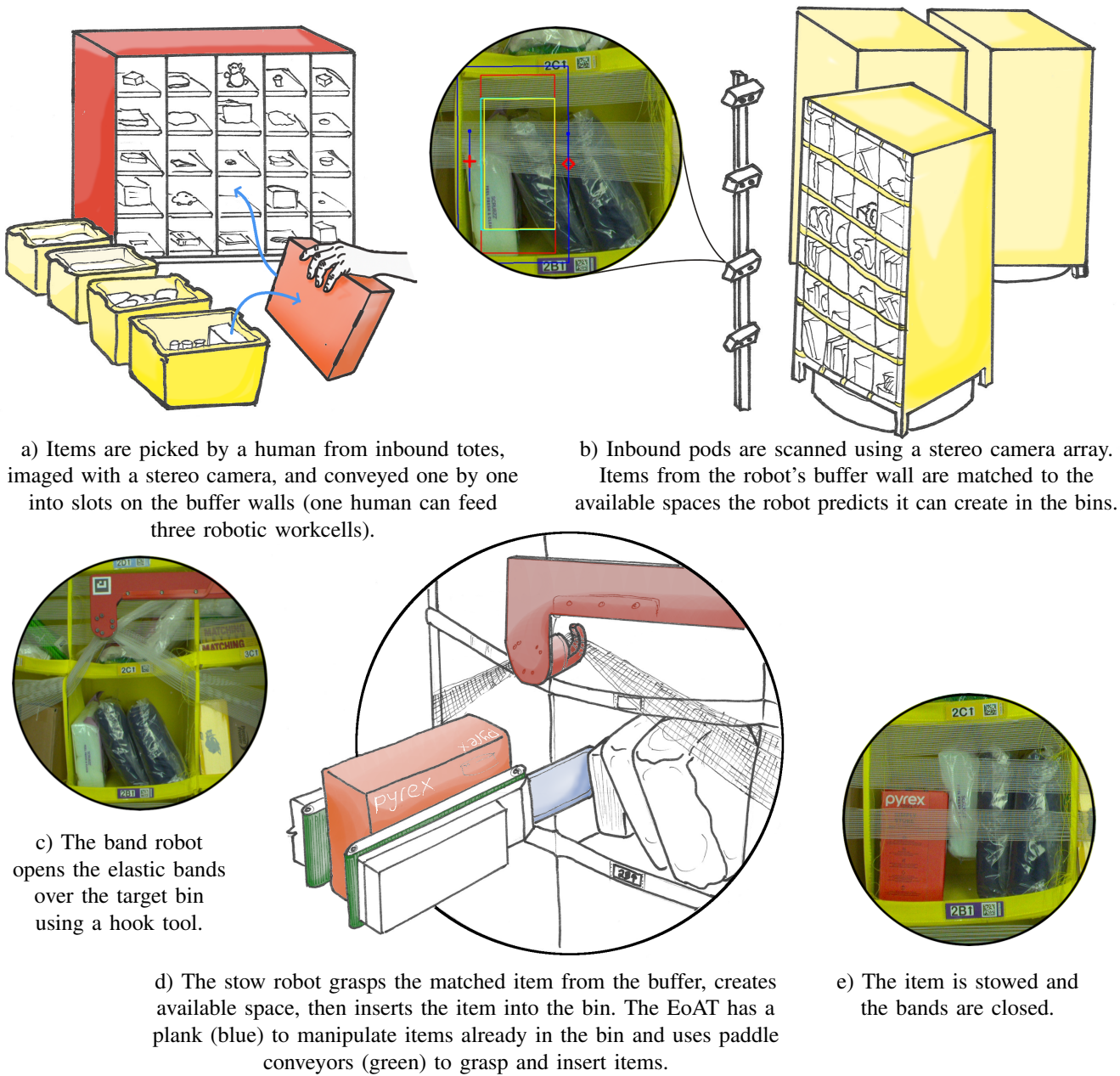
\begin{figure*}[!ht]
\centering
\begin{overpic}[width=\textwidth]{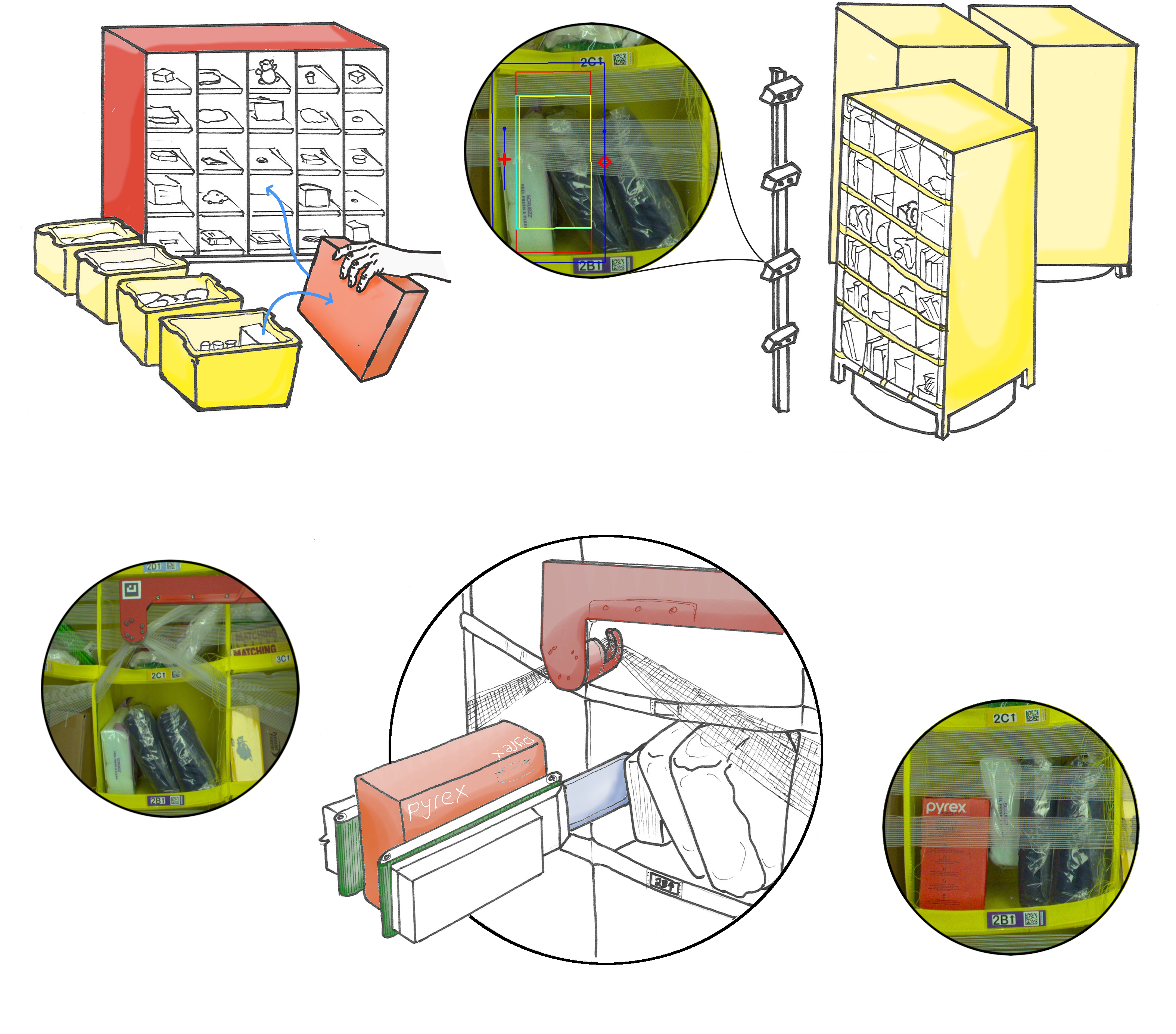}
    \put(3,49){\parbox[t]{8cm}{\centering a) Items are picked by a human from inbound totes, imaged with a stereo camera, and conveyed one by one into slots on the buffer walls (one human can feed three robotic workcells).}}
    \put(52,49){\parbox[t]{8.5cm}{\centering b) Inbound pods are scanned using a stereo camera array. Items from the robot's buffer wall are matched to the available spaces the robot predicts it can create in the bins.}}
    \put(3,15){\parbox[t]{3.5cm}{\centering c) The band robot opens the elastic bands over the target bin using a hook tool.}}
    \put(15,3){\parbox[t]{10cm}{\centering d) The stow robot grasps the matched item from the buffer, creates available space, then inserts the item into the bin. The EoAT has a plank (blue) to manipulate items already in the bin and uses paddle conveyors (green) to grasp and insert items.}}
    \put(74,3){\parbox[t]{4cm}{\centering e) The item is stowed and the bands are closed.}}
\end{overpic}
\vspace{13pt}  
\caption{In the Robotic Stow Process, a human checks inbound items for quality and feeds three robotic workcells. The workcells store items in buffer, then match and store items to inbound pods using a set of manipulation behaviors.}
\label{fig:StowProcess}
\end{figure*}

The robotic stow process, shown in Figure \ref{fig:StowProcess}, begins with items arriving in plastic totes. A human inductor checks each item for quality (e.g. ensure there is no pre-existing damage) and robot eligibility (e.g. no liquids) before placing it into the system. Conventional material handling equipment transports the singulated items to one of several interconnected robotic workcells. Each workcell stores singulated items in a buffer wall that can be accessed immediately when the robot requests a specific item. Having a diversity of item shapes and sizes in the buffer increases the likelihood of finding a stow opportunity. Currently each buffer wall has 32 slots and each human inductor feeds three robotic workcells.

When a pod arrives at a robotic workcell, it is imaged by a set of passive stereo cameras. The perception system fuses a learned depth model and segmentation model to create an orthographic map of each bin. These models are trained to be invariant to the occluding elastic bands. Having an accurate world model is not sufficient, though. This orthographic map is then used by space estimation algorithms to predict the available space that could be created after objects in the bin are reconfigured. 

A bin-item match algorithm uses these space estimates along with 3D information about each item in the buffer to match items to bins. This algorithm predicts the risk of each match along with the expected execution time to optimize the stow rate. It also selects one of a canonical set of behaviors to execute the stow. 

When a match has been made, the item is retrieved from the buffer and vended to a reorientation system. Items are rotated before being grasped by the robot so that they will fit tightly into bins and so that preferred orientations (e.g. book titles facing out) can be respected. The robot end effector (or End-of-Arm-Tool - EoAT) grasps items between two paddles, each with an active, built-in conveyor belt. This allows items to be fed into the grasp via conveyor transfer rather than 'picked'. The EoAT also has a thin, extendable plank of aluminum which is used for in-bin item manipulation. This provides a known engineering surface through which contact is made with the world. The geometry of this plank also minimizes the volume of hardware inserted into the bin, helping to maximize the fullness of the bins. 

Each stow behavior uses kinesthetic feedback and high-rate force/torque sensing to manipulate items in the bin and insert items into the created spaces. The set of discrete behaviors provides a variety of strategies for different bin configurations and item types. For example, a lateral sweep motion can consolidate a mixture of bagged apparel just enough to fit a thin box into the bin while a flip up motion can take a single book that has fallen over and lean it against the sidewall of the bin to create a large space for a basketball. Using each bin's orthographic map, a set of affordances are computed to choose specific behaviors to execute in a bin. 

As the EoAT grasps the selected item from the buffer and reorientation system, a second manipulator is used to open the elastic bands covering the bin. This manipulator uses the same orthographic map to decide how to grasp the bands and avoid touching objects in the bin.

After each stow is completed, space estimates are updated using kinesthetic information and new images of the bin. Space estimation, match, and behavior execution can occur largely in parallel due to the number of bin choices, thus cycles can overlap so that the robot is never waiting for the next requested item to arrive from the buffer and reorientation system. 

Key metrics for the process include the stow rate in units per hour and the density achieved in the bins, often measured in volumetric occupancy or gross cubic utilization. The number of items stowed per pod face is important because of its impact on stow rate. After stowing all the items into a pod that have matched (on average 8), the mobile robot takes the pod back into the storage field while another pod is brought into position. This pod transition time takes approximately 6 seconds, so it is beneficial to stow as many items into a pod face as possible to maximize rate. Especially damaging to the units per hour is when no items are stowed into a pod -- this is called a no-stow-turnaway -- as it consumes at least 12 seconds. The system must also minimize defects like dropped items (amnesty), damaged items, and mismatches between the physical world and virtual inventory management database. Finally, the system must run autonomously without interruptions that require human intervention (e.g. to clear a jam, restart a subsystem, or jog a robot) and without failures that limit the life of the system or require downtime for repairs.

\section{Contributions}

This paper describes the key robot morphology choices and hardware designs in Section \ref{section:hardware}, and perception, motion, and task planning approaches in Sections \ref{section:Perception}-\ref{sec:match}. We then report results from a deployment in an e-commerce warehouse, analyzing a recent batch of 100,000 stows attempted by the system in Section \ref{section:results}. 

In addition to presenting the overall robotic stow system, this work includes several novel contributions: 1) A non-anthropomorphic pinch gripper design that uses an extendable plank to simplify the manipulation of items within the bin and conveyors on each `finger' to place items into confined spaces and reduce item damage while maintaining a stable item pose; 2) A perception system that can predict space through semi-transparent elastic bands. This system extends CRE-Stereo \cite{Li2022CVPR} to see though occluding bands and combines it with instance segmentation, heuristics and models learned from experience; 3) A 'risk-aware' bin packing planner that combines a heuristic similar to best-fit \cite{Lodi2002AppliedMath} with learned models that predict the probability of behavior success conditioned on the perceived bin state. 4) A set of behaviors that leverage the unique robot morphology to make space in the bin and detect off nominal conditions using kinesthetic feedback. Finally real world performance results and important error cases are discussed that may help motivate adjacent research.




\section{Hardware Design}\label{section:hardware}
\subsection{Manipulation Topologies}
The stow process requires multiple manipulation tasks -- item singulation from totes, pre-stow item reorientation, elastic band manipulation, manipulation of items already in a bin to create space, and stowing the item. Our solution biases towards specialized manipulators rather than fewer, generalized robots. This enables parallelization and higher reliability for each task.

\begin{figure}[t]
\centering
\includegraphics[width=1.0\linewidth]{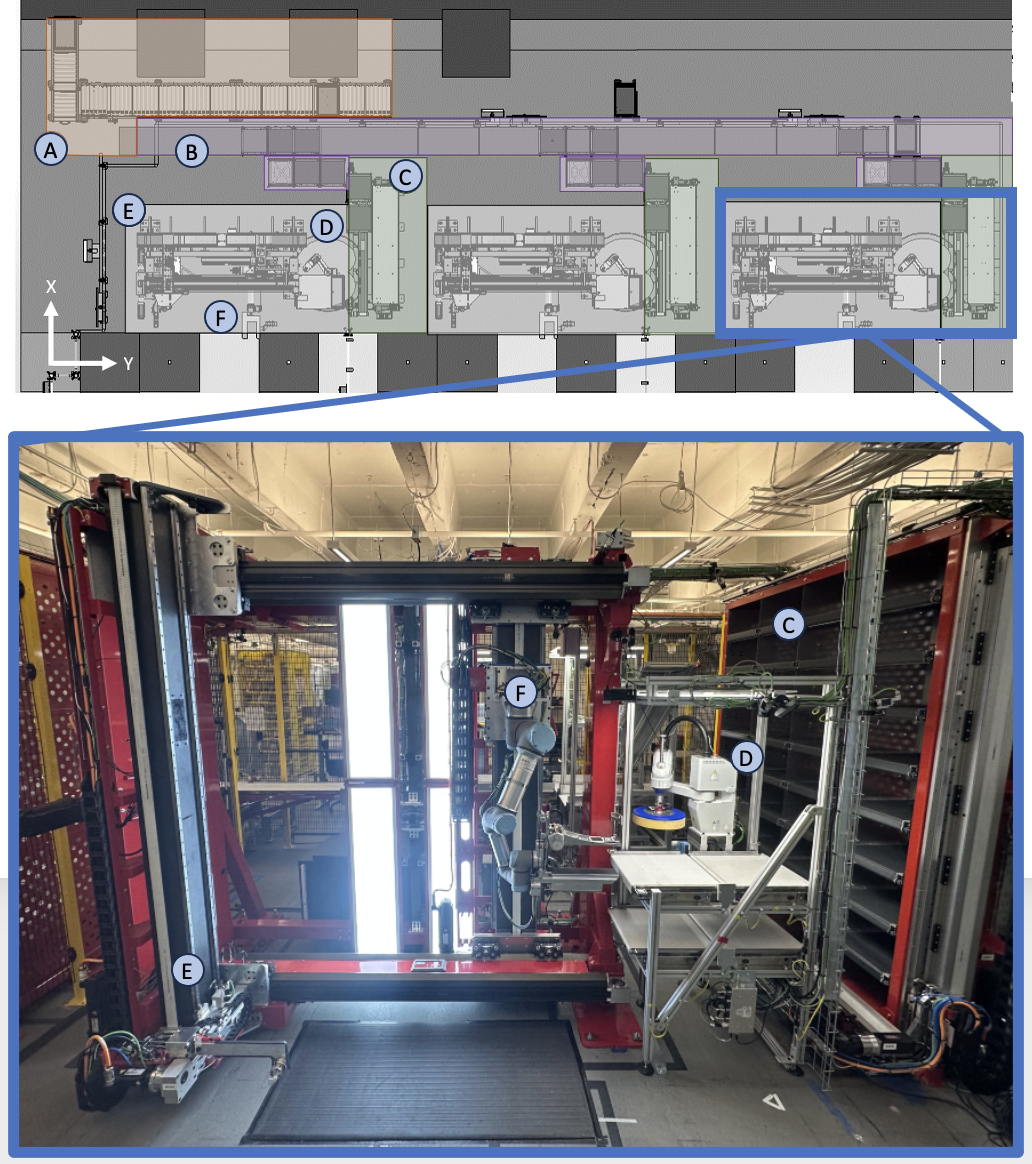}
\caption{Top: Plan view of system with A) induct station, B) transfer conveyance, C) Item Buffer, D) Reorientation, E) Band Manipulator, F) Stow Manipulator.  Bottom: Robot station where items are vended by C) Item Buffer then reoriented by D) SCARA robot before being grasped and stowed by F) robot arm on gross positioning gantry. E) linear gantry robot opens elastic bands.}
\label{fig:Alpha2}
\end{figure}

Figure \ref{fig:Alpha2}--top shows a plan view of the robotic system. Items flow from the human induct station (A) to one of three identical robotic systems via conveyance (B). The items are buffered (C) until they are matched to a stow opportunity. Once matched, they are reoriented (D) before being fed to the Stow robot (F). In parallel to the grasp, the band separation robot (E) pulls the elastic bands out of the way to enable the stow.

Item singulation from totes is currently performed by a human to leverage their ability to quality check items (e.g. for expiration date violation, open box, damaged item) and problem solve. Items are placed on the conveyance with their shortest axis aligned to the gravity vector. This pose maximizes item stability during conveyance throughout the system. 

Pre-stow item reorientation is performed by a SCARA robot that rotates the item around the gravity vector on a low-friction conveyor belt using a foam disc, shown in Figure \ref{fig:Infeed}. With a relatively loose tolerance requirement of a few degrees and forgiving success criteria of 98\% (with an opportunity to reattempt), this system is low cost and avoids generating defects that might occur while picking up the item to perform the reorientation. Items that are not successfully reoriented can be easily restored to the item buffer or recycled out of the system for human stowing.

Elastic band manipulation is performed by a three-axis linear gantry robot designed to match the tall but narrow workspace of a pod. This architecture also minimizes the collision volume for the stow robot's behaviors by using a cantilevered end of arm tool.

Finally, manipulation of items already in the bin and the item insertion task are both performed by a single serial kinematic chain. These two functions were paired because once bin manipulation is complete, the gripper is in the preferred location for stowing the item. A six degree of freedom robot arm is grossly positioned by a two-axis gantry as shown in Figure \ref{fig:Alpha2}--bottom. This 2+6 degrees of freedom enables the robot arm to have high manipulability (good condition number) no matter which bin it is stowing into. In contrast, a statically mounted robot with a larger reach would be near a singularity for upper and lower rows of bins and also have a large collision volume for the robot elbow to contend with in the tight workspace available. Ceiling height limitations and floor loading criteria also restrict how large of a robot can be deployed to stow station locations.

\subsection{Item Buffer}
The item buffer currently holds 32 items in four columns of eight slots. The buffer wall itself is completely passive with no motors or sensors. A vertical-horizontal gantry that carries two active conveyors traverses in front of the buffer wall to store and retrieve items. In a typical cycle, an item is taken from the transfer conveyance onto the first gantry conveyor. A matched item is then pulled from the buffer wall onto the second conveyor. The third step in the sequence is to place the transfer item into the vacated buffer slot. Finally, the matched item is vended to the reorientation system. This cycle can be completed in eight seconds on average, ensuring down stream manipulations are never starved of items. Exception handling sequences exist to reprocess items that fail reorientation or that the robot is unable to stow (e.g. when the predicted amount of space is not created). Faster sequences are also possible to quickly fill the buffer walls before stowing begins or to vend items consecutively to the robot when there are no items queued in the transfer conveyance.


\subsection{Stow Station}
Because the stow robot and elastic band robot never cross each other in the horizontal direction, a shared structure can be used. The elastic band manipulator uses two rails (top and bottom) for its horizontal motion, synchronized through torque tubes so that a single motor can be used.  A vertical column belt drive provides access from the top of the pod face to the bottom, and another belt drive moves the band separation hook into contact with the pod face for manipulation and retracts it for free space motion. Band separation motions require less than three seconds to execute, and motions are coordinated across an EtherCAT field bus.

The base of the robot arm is positioned by a horizontal-vertical gantry. The horizontal gantry architecture is copied from the band separation robot and a belt drive is used for the vertical actuator. This 2+6 degree-of-freedom kinematics enables full access to the pod face and high manipulability in each task. In the first position of a stow cycle, shown in Figure \ref{fig:Alpha2}, the robot grasps an item from the reorientation stage. In reality, this grasp is a simple conveyor-to-conveyor transfer due to the robot's gripper design (described next). This grasp pose is designed so that the camera tower is unobstructed and the system can update its understanding of the pod face scene. 

Once an item is grasped, the gantry transports the robot arm to a position in front of the intended bin. Simultaneously, the robot arm moves to its pre-stow pose. The accelerations from the gantry motion require intelligently planning this coordinated 8-DOF motion to avoid triggering protective stops. The task-space stow motions are performed in the local frame of the bin, executed exclusively by the robot arm and gripper. 

\subsection{End of Arm Tool Designs}

\begin{figure}
\centering
\includegraphics[width=1.0\linewidth]{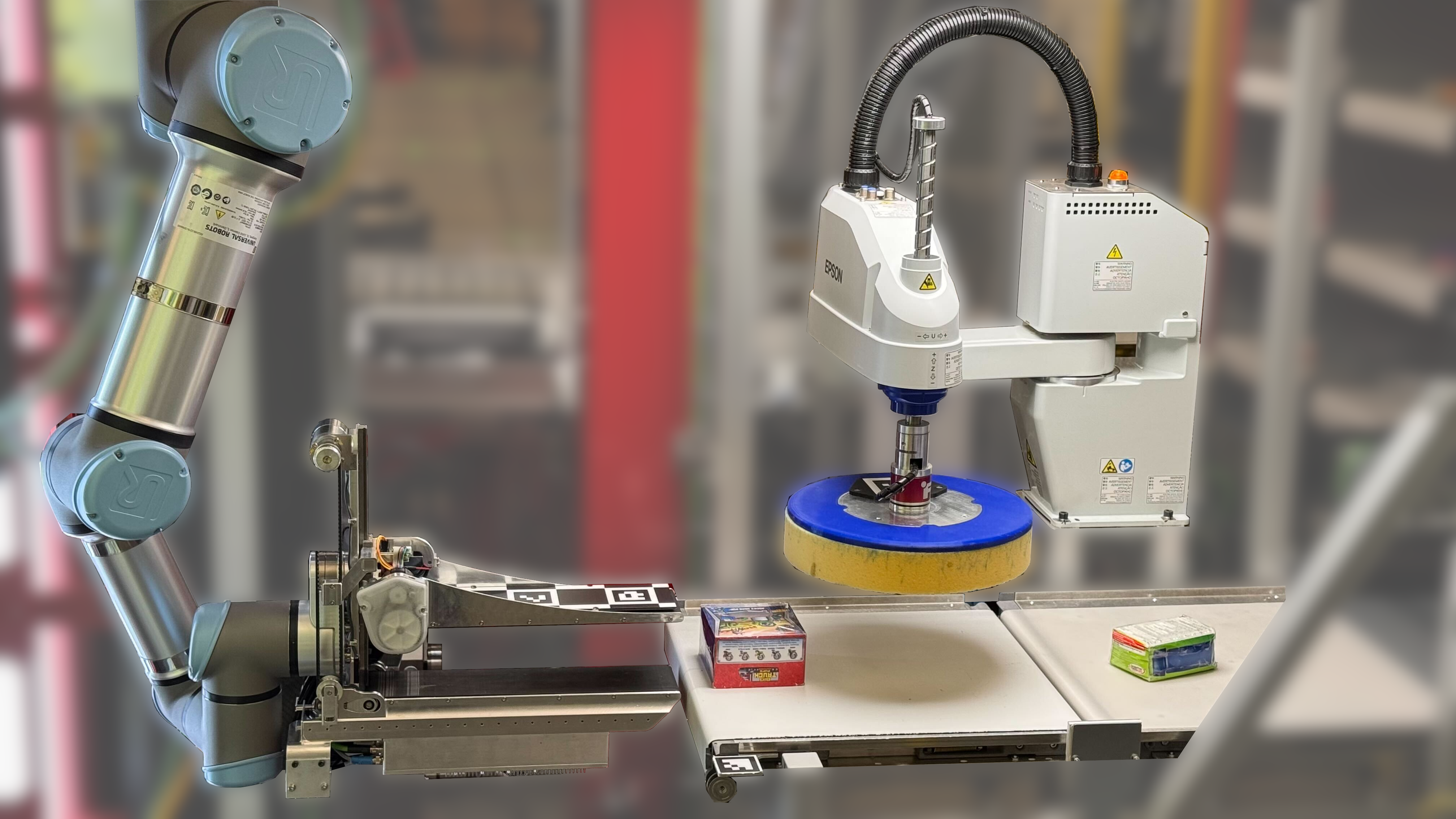}
\caption{Orientation EoAT and Hand-off: A SCARA robot with a foam disk is used to orientate items before they are conveyed into the robot EoAT.}
\label{fig:Infeed}
\end{figure}

\subsubsection{Orientation EoAT}
Item reorientation rotates items such that the long axis of the item is parallel to the direction of conveyor travel, which maximizes the fullness of bins in almost all cases. This prevents any re-grasping of the item or in-hand manipulation requirements for the Robot EOAT. In a few exception cases, the long axis of the item is aligned orthogonal to the direction of conveyor travel. Examples include books with their titles facing outwards (to facilitate quick identification when they are later picked from the bin), behaviors that stack items rather than placing them side-by-side, and a small minority of bins that are taller than they are deep. Reorientation is achieved with a SCARA robot arm applying an 80 N downward force on the item resting on a low-friction conveyor belt surface through a 30-cm diameter, 5-cm thick polyurethane foam disk (see Figure \ref{fig:Infeed}).  The foam disk is then rotated, rotating the item along with it. For more fragile items, lower forces can be used to prevent item damage. Before and after images of the item determine the rotation commands and confirm success or failure of the attempt.

\begin{figure*}[ht]
    \centering
    \subfloat[Robot EoAT]{%
        \includegraphics[width=0.65\textwidth]{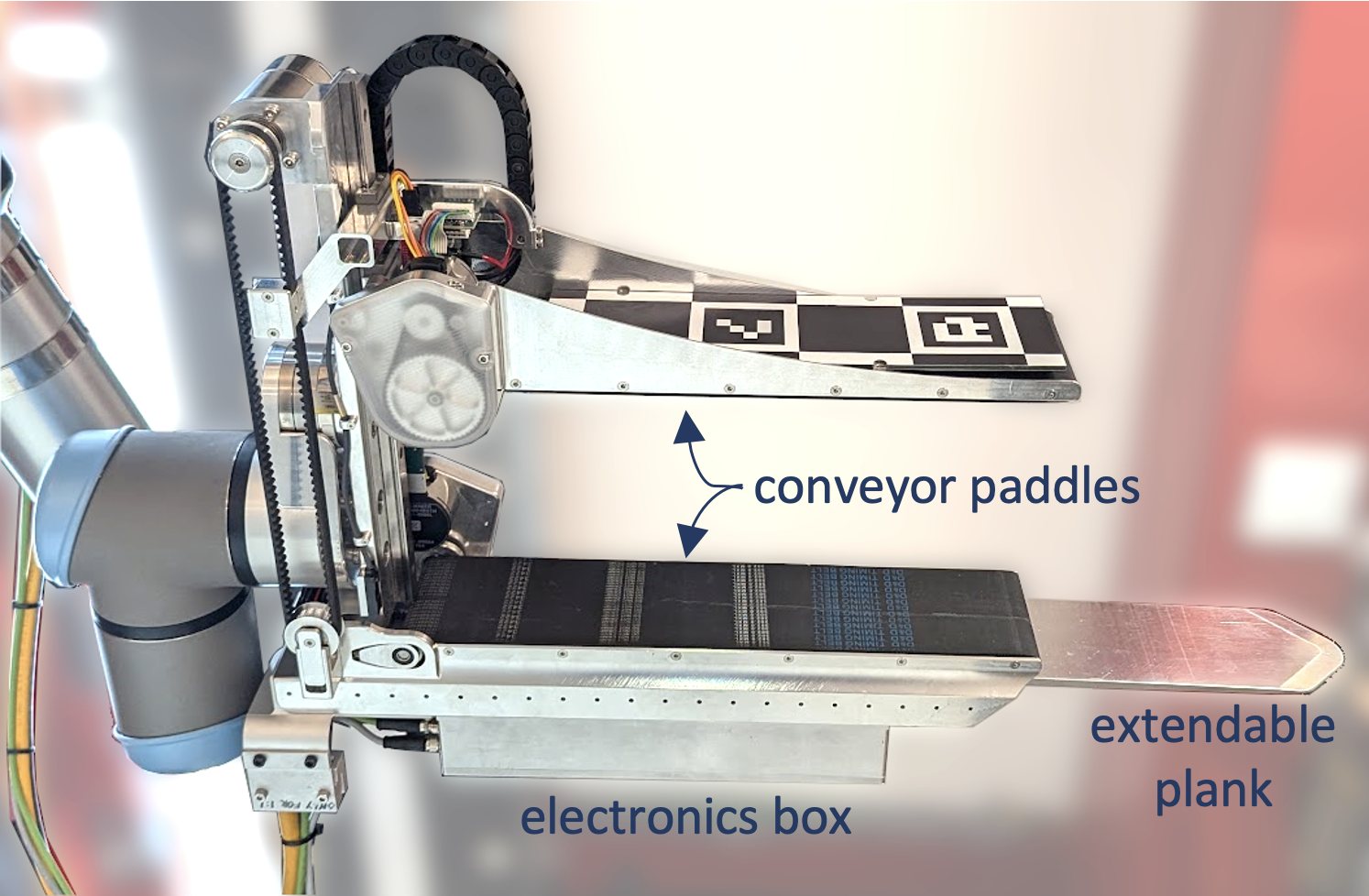}%
        \label{fig:EoAT}%
    }
    \hfill
    \subfloat[Band EoAT]{%
        \includegraphics[width=0.30\textwidth]{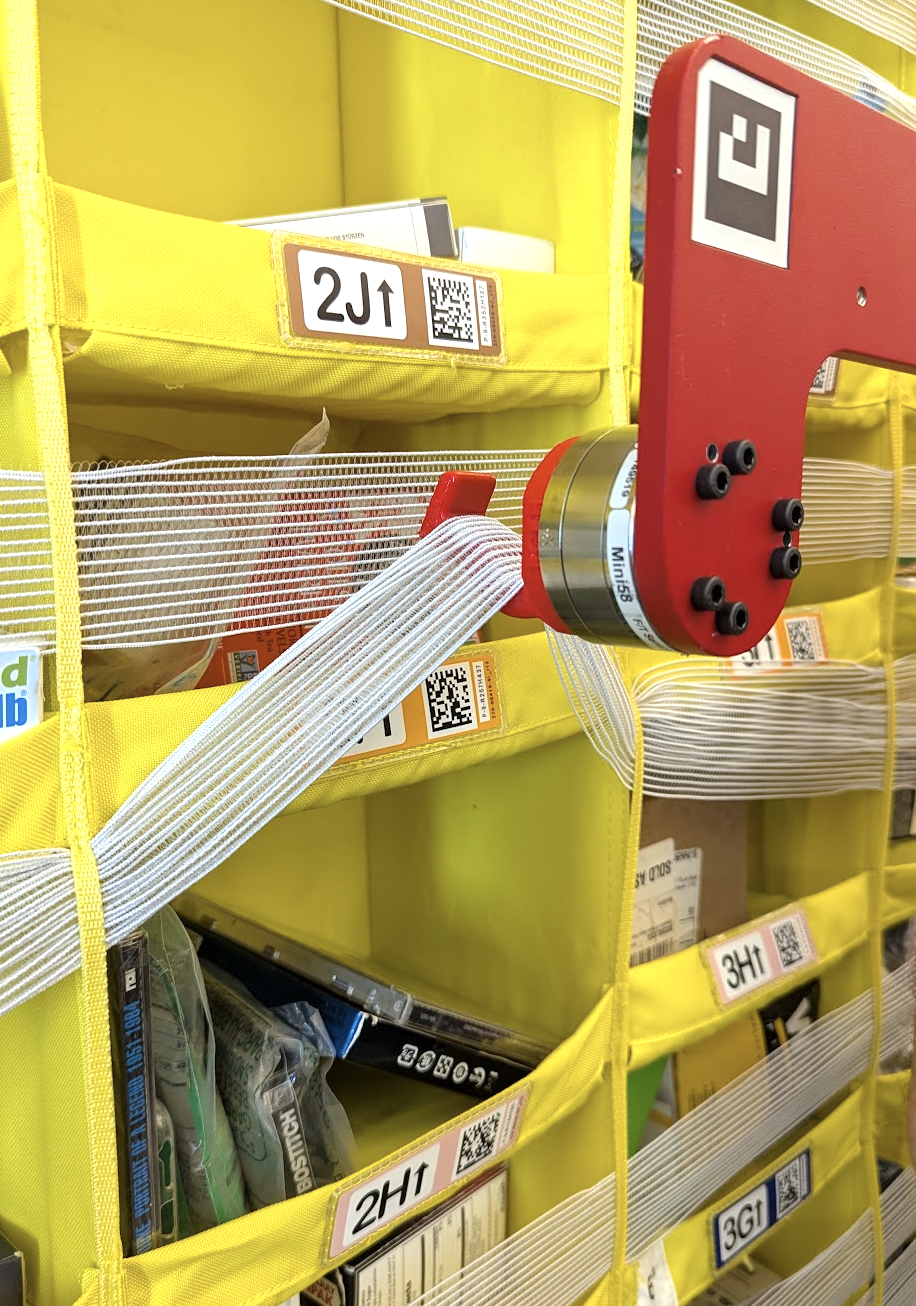}%
        \label{fig:second}%
    }
    \caption{a) The robot EoAT uses parallel jaws with built-in conveyors. With this strategy, items are fed into the jaws by simple conveyor transfer and inserted into bins without requiring arm motions A retractable aluminum plank is used to manipulate items already in the bin to create available space to stow. b) The band EOAT uses a compound hook to grasp and pull the elastic bands out of the way for stowing. With a cantilevered profile, collision volumes are minimized, and with a six-axis force-torque sensor, contact can be made safely with the pod bin.}
    \label{fig:both}
\end{figure*}

\subsubsection{Robot EoAT}
A large parallel-jaw gripper is sufficient for successfully achieving force closure on most items in a fulfillment center. Adding conveyors to the surfaces of the jaws facilitates a reliable item transfer (grasp) from the reorientation conveyor into the EoAT by advancing both conveyors at the same rate. The conveyors also facilitate a controlled insertion of the item into the bin without requiring the jaws to be inserted into the bin, which would negatively impact achievable fullness. With this method of item insertion, the pose of the item is maintained throughout the insertion motion as well, minimizing unintended collisions with the environment. If a collision does prevent a successful stow, this can be detected (Section \ref{sec:insertion-exception-handling}), and the jaw-conveyors are run in reverse to pull the item back into the gripper so that it can be recycled or restored to the item buffer. The jaws regulate closing force using a single degree of freedom load cell to measures grasp force, and limit switches are used to home the tool.

Bin manipulation is achieved via a tapered plank of aluminum. A six-axis force-torque sensor is integrated at the proximal end of the aluminum, providing information about the contact loads between the plank and the environment. This includes the pod structure, bin walls, and items already in the bin. A distributed motor control approach was chosen to minimize the dress pack on the robot. A custom circuit board controls the four actuators on the EOAT and processes all of the sensing inputs. Only EtherCAT and power cables are required. 

\subsubsection{Band EoAT}
Each bin has several 5-cm wide elastic bands spanning the face. The band position is not consistent from bin to bin nor from stow to stow for the same bin. Items in the bin may be pressed up against the bands. An aluminum hook geometry is used to slide under the bands and in between items in order to pull the bands up and away from the bin face. A six-axis force-torque sensor enables these motions to be planned and controlled using contact information. The tool is cantilevered from the band robot to minimize the collision volume near the active bin.

\subsection{Cameras and Fiducials}
The pod scene is illuminated by 6500 K LED panels with approximately 22k lumens coming from approximately 2 m$^2$ of surface area (see Fig. \ref{fig:Alpha2}).  Images are captured by four pairs of stereo cameras with a 100-mm intraoccular distance and a 0.8 m vertical spacing between pairs. Fiducials are integrated into the band separator and robot EoATs to enable extrinsic calibration of the EoATs to the camera frames.

Conveyor scenes are imaged by stereo cameras positioned above the first conveyor at the induct station, the item reorientation conveyor, and the item infeed conveyor for each robot workcell. Identical hardware supports the use of a common model at these locations.

\section{Perception}\label{section:Perception}

The robot uses the vertical array of stereo cameras to image each pod when it arrives. The perception system then creates a state representation for each bin, which includes the bin walls and lip, items within the bin, and the bands which overlay it. The bands (shown in Figure \ref{fig:StoragePod}) are an elastic mesh comprised of thin white strands, which creates a translucent effect occluding objects behind them, through which the perception system must infer item state. The learned bin representation is then used to estimate the available space in each bin and later in deciding what behaviors to execute to realize that space. 

\subsection{Learned Bin Representation}\label{sec:bin}
\begin{figure*}[t]
\centering
\subfloat[Synthetic RGB image of pod with bands.]{\includegraphics[width=1.2in]{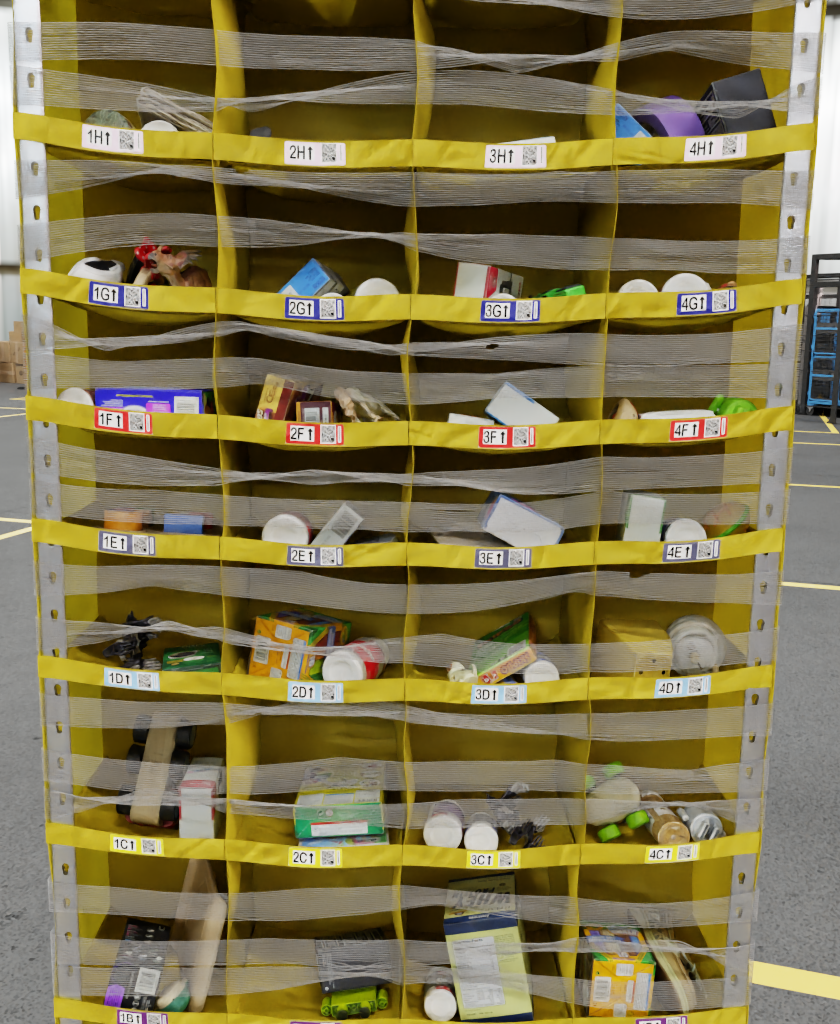}\label{fig:SyntheticExamples-a}}
\hfill
\subfloat[Instance segmentation of bins in the pod.]{\includegraphics[width=1.2in]{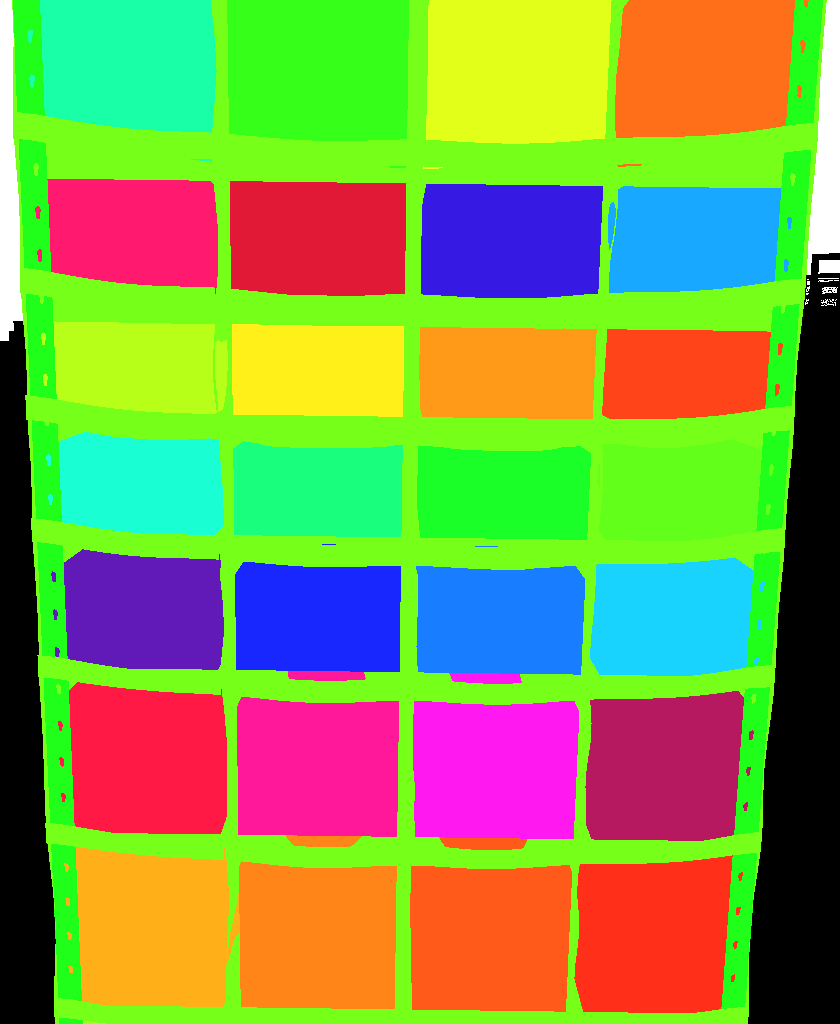}\label{fig:SyntheticExamples-b}}
\hfill
\subfloat[Semantic segmentation of bands in each bin.]{\includegraphics[width=1.2in]{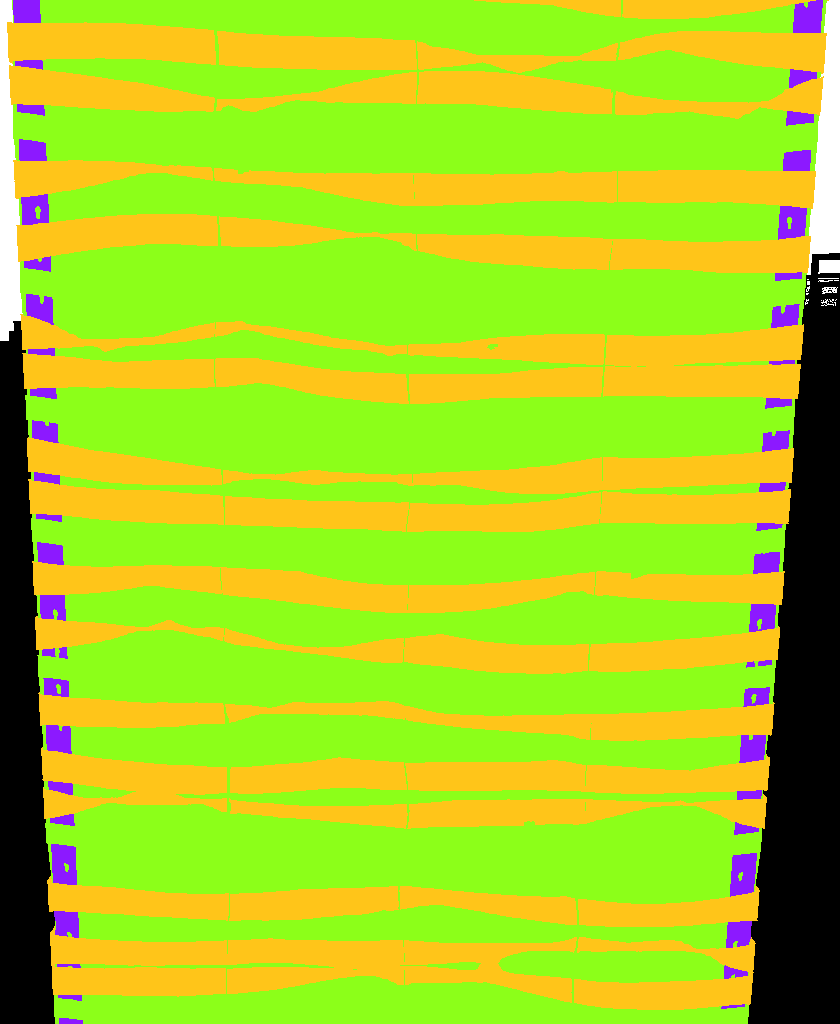}\label{fig:SyntheticExamples-c}}
\hfill
\subfloat[Instance segmentation of items in bins while seeing through bands.]{\includegraphics[width=1.2in]{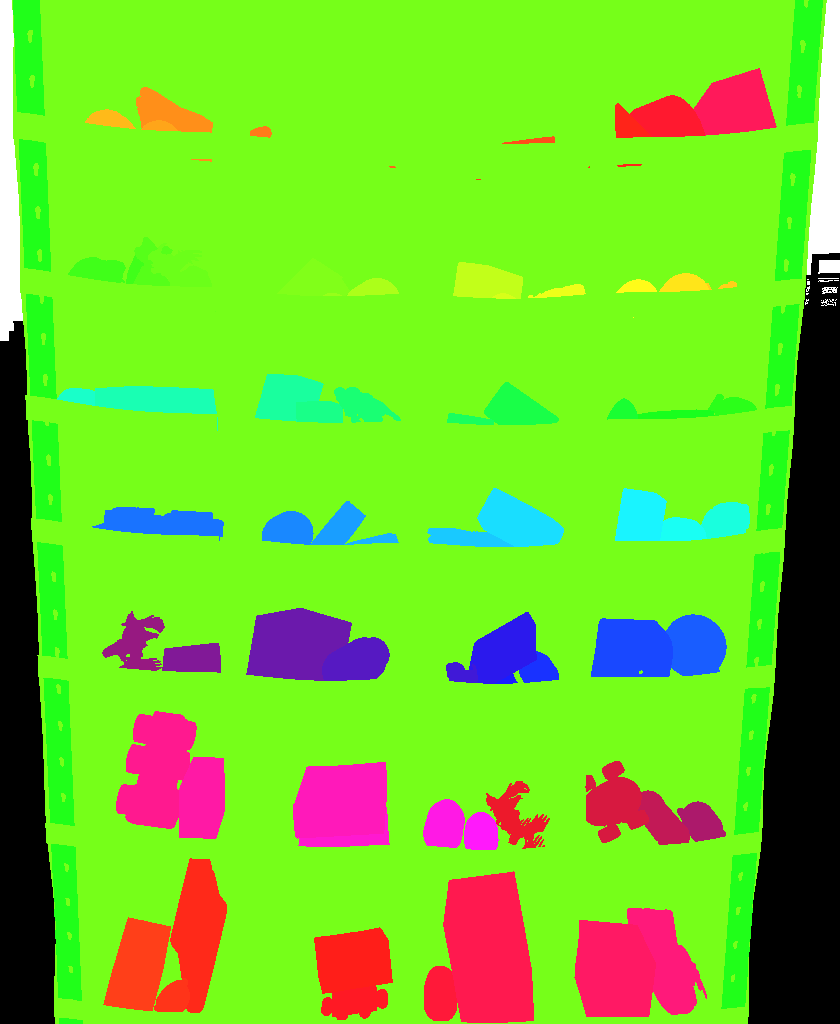}\label{fig:SyntheticExamples-d}}
\hfill
\subfloat[Depth maps of the bins and items while seeing through bands.]{\includegraphics[width=1.2in]{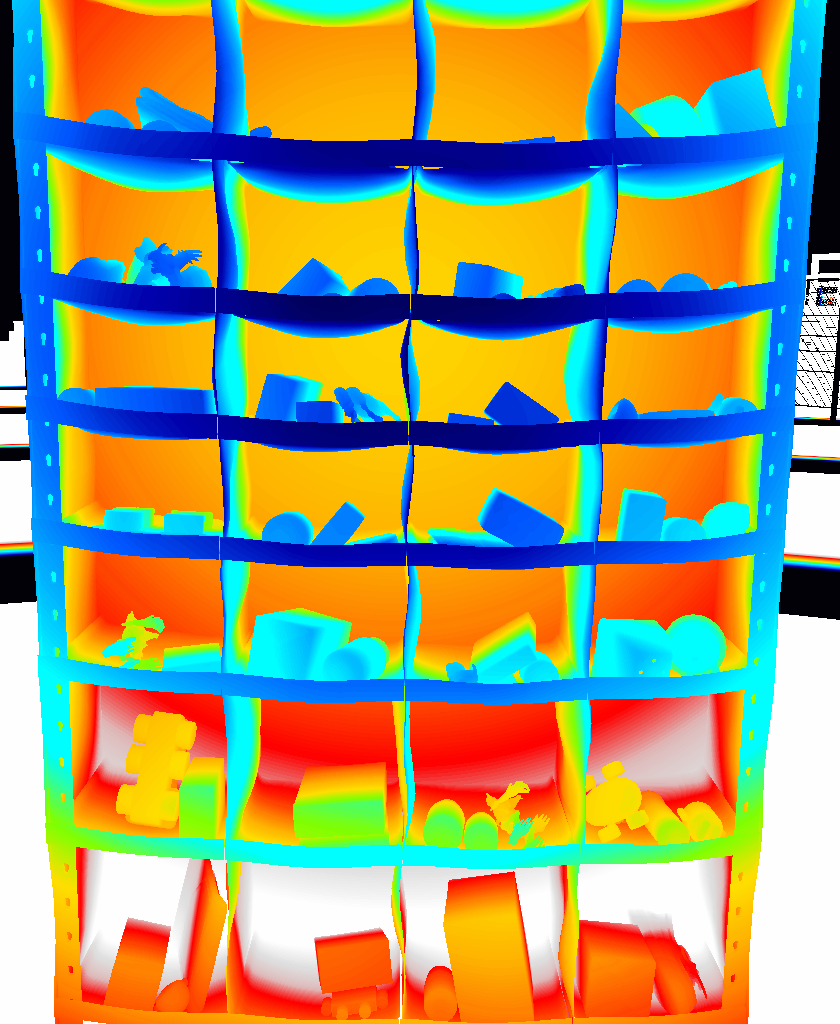}\label{fig:SyntheticExamples-e}}
\caption{Synthetic images of pods along with their segmentation and depth maps that see through partially occluding translucent bands in bins. This is used in conjunction with non-synthetic images to train the models for learned bin representations.}
\label{fig:SyntheticExamples}
\end{figure*}

The state of each bin is represented as a layered orthographic map, referred to here as a \emph{multi-mask}. Individual layers of this multi-mask are computed using the output of learned depth and segmentation masks. The bands can be detected using standard semantic segmentation methods, whereas items and bin layers use learned depth and segmentation models trained to see through occluding bands using simulation. Each of these models have been trained using supervised labels with a combination of synthetic images (see Figure \ref{fig:SyntheticExamples}) for pre-training and manually annotated non-synthetic images captured by cameras in the workcell for fine-tuning. This section discusses their respective model architectures. 

\subsubsection{Bin Instance Segmentation} \label{section:bin-instance-segmentation}
This model operates on images of pods and detects instance segmentation masks for each bin, which is then used to extract crops of individual bins. An illustration of the input image of the pod and the output bin instance segmentation is shown in Figures \ref{fig:SyntheticExamples} (a) and (b), respectively. The remaining segmentation and depth models discussed next in this section operate subsequently on the individual images of bins extracted using the bin instance masks. Given that the diversity in pods and bin shapes is typically much more limited compared to items, this model utilizes the lightweight architecture of RTMDet-s \cite{Lyu2022arXiv}. When ingesting RGB images of pods with resolution 640 $\times$ 768, the model operates at an inference latency of around 25ms on a GeForce RTX 4080 GPU to detect up to 50 bins in pods.

\subsubsection{Band Semantic Segmentation}
A semantic segmentation model is used to predict binary masks representing the translucent bands. An illustration of the semantic masks for the bands is included in Figure \ref{fig:SyntheticExamples-c}. This semantic segmentation utilizes the architecture of Fully Convolutional Networks (FCN) \cite{Long2015CVPR} with a ResNet \cite{He2016CVPR} backbone. When ingesting RGB images of bins with resolution 256 $\times$ 256, the model operates at an inference latency of around 5ms on a GeForce RTX 4080 GPU.

\subsubsection{Item Instance Segmentation (Seeing through bands)} \label{section:item-instance-segmentation}
The item instance segmentation model ingests each bin image and predicts instance segmentation masks for each individual item within the bin. An illustration of the instance masks for each item in bins, while seeing through bands, is included in Figure \ref{fig:SyntheticExamples-d}. This model leverages the architecture of Mask R-CNN \cite{He2017ICCV} with a ResNet \cite{He2016CVPR} backbone and is trained using RGB images of bins with and without bands as input, while item instance masks without bands are used as its target output. When ingesting RGB images of bins with resolution 256 $\times$ 256, the model operates at an inference latency of around 50ms on a GeForce RTX 4080 GPU.

\subsubsection{Depth Prediction (Seeing through bands)}
\label{section:depth-prediction}
The depth prediction model takes a pair of rectified primary and secondary stereo images for each bin as its input and predicts a disparity map $\mathcal{\tilde{D}}$ for the pixel correspondences between them as the output. As we'll discuss in this section, any offset $\tilde{d}$ between the coordinates used to extract the bin images from the primary/secondary stereo images is added back to this disparity map to get the final disparity map $\mathcal{D} = \mathcal{\tilde{D}} + \tilde{d}$. Based on the principles of stereo vision, the final disparity map $\mathcal{D}$ is then converted to a depth map $\mathcal{X}$ for the primary view using scalar multiplications with the camara's focal distance $f$ and the baseline distance $b$ between the stereo pair as $\mathcal{X} = f \times b / \mathcal{D}$ \cite{Hartley2004Cambridge}. This depth map represents the distance of the items and bins from the camera along its optical axis.  

This depth prediction model differs from other regular forms of learned stereo in two key aspects: (1) the model is trained to see through the translucent bands partially occluding the bins and estimate the depth maps as if the bands were invisible in the scene; (2) the model predicts depth for a specific region of interest (representing a given bin) in the pair of stereo images of the pod. Meanwhile, the motivation for (2) is to maximize the pixel resolution representing a given bin in the input image pairs provided to the depth model, and it also allows updating of specific bins on demand at lower latency. For (2), we determine the image coordinates for extracting tight crops of each bin in the primary pod image based on the bin instance masks discussed in Section \ref{section:bin-instance-segmentation}. The corresponding image coordinates for tight crops of the respective same bins in the secondary pod image are extracted based on the expected horizontal offset $\tilde{d} = f \times b/\tilde{x}$, where $\tilde{x}$ represents an estimate of the pod's distance from the camera. This value of $\tilde{x}$ is estimated based on algorithmically localizing the pod based on stereo vision. As aforementioned, this offset $\tilde{d}$ in the image coordinates is added back to the output disparity map $\mathcal{\tilde{D}}$ before it is converted to the depth map $\mathcal{X}$. This approach allows for tight bin crops from both the primary and secondary views, to be sent to the learned stereo algorithm, maximizing pixel resolution and increasing overlap in both crops.

The depth model leverages the architecture of CRE-Stereo \cite{Li2022CVPR} and is trained using RGB images of bins with and without bands as input, while disparity maps without bands are used as its target output. In comparison to a monocular approach to predict the depth of objects partially occluded by translucent surfaces \cite{Dai2025ICRA}, this stereo-based approach achieves a reduction of over 45\% in $L_1$ depth error, decreasing it from 2.14 cm to 1.16 cm, in A/B evaluations on identical datasets.  When ingesting RGB images of bins with resolution 256 $\times$ 256, the model operates at an inference latency of around 80ms on a GeForce RTX 4080 GPU.

\subsubsection{Perspective-Corrected and Unified Multi-Mask for Bin Representation}
\label{section:multi-mask}
To facilitate downstream processing, we construct a unified representation for each bin, referred to as a multi-mask. This representation integrates the outputs of the aforementioned models by applying an orthographic reprojection to the bin segmentation, band segmentation, item instance segmentation, and depth map. The orthographic reprojection, made possible by the depth map, aims to correct for perspective distortion by simulating a frontal view of each bin as captured from a virtual camera positioned at an infinite distance along the optical axis, centered at the 2D centroid of the bin. The resulting orthographic projections of the model outputs are then concatenated as separate channels to form the multi-mask representation of each bin. This is then used for downstream applications such as the estimation of free-space that remains within each bin, as discussed in the next sub-section below. Section \ref{sec:behaviors} and Figure \ref{fig:affordance_generation} subsequently describe using this multi-mask to generate and select behaviors.

\subsection{Estimation of Free Space in Bins}\label{sec:space_estimation}

Accurate prediction of the free space that can be created within each bin is critical for matching items in the buffer to suitable bins within the pod. This estimation also informs the selection of behaviors, which aim to create the required space and execute the planned placements. The components of free space estimation, item-to-bin matching, and behavior selection are hence tightly coupled within the end-to-end system.  The comparison of predicted space and the actual space created by executed behaviors, as measured by force sensors on the robotic arm, directly provides feedback that can continually improve the accuracy of free-space prediction over time.

To ensure that bins remain accessible for subsequent picking operations, the stowage process adheres to a set of etiquette conventions. This includes maintaining visibility of all items from the frontal view of the bin, favoring placements along the horizontal axis spanning the left and right edges of the bin, and avoiding placements that obstruct visibility along the depth axis. As a result, free-space prediction emphasizes the horizontal axis to align with these ergonomic and operational constraints.

The predicted free space can be broadly categorized into two types: (1) Directly usable space, which can accommodate new items without requiring any manipulation of existing contents; (2) Manipulable space, which can be created by using the EoATs extendable plank to move and compress items in the bin. Category (2) can be further subdivided into two modeling approaches: (a) assuming that items are rigid, non-compressible bodies and (b) accounting for item-specific compressibility. For category (1), the system computes the width of the largest contiguous rectangular region in the bin’s multi-mask representation that does not intersect with any item occupancy masks. Here, item occupancy masks refer to the instance segmentation masks of items, while only considering pixels in the mask between the front face and the mid-point along the bin's depth. This design choice reduces the risk of underestimating space due to item orientations with non-zero yaw angles. In category (2)(a), the system fits oriented 2D bounding boxes around segmented items and estimates each item’s width based on its bounding box. The total free space is then calculated as the difference between the bin’s width (derived from the bin instance mask) and the sum of the individual item widths. For category (2)(b), if the most recent stow in a bin was performed by the robotic workcell—as opposed to a human stower—additional historical information is available. This includes kinesthetic position and force sensor traces, which capture the free space created, and the gripper width, which measures the width of the previously stowed item. The difference between these two quantities implicitly captures the bins remaining space and directly accounts for item compressibility. This is used as a refined estimate of the available free space that accounts for item deformation under compression. When available, the system prioritizes the compressibility-informed estimate from category (2)(b) as its final free-space prediction. Else, it selects the maximum value between the estimates from categories (1) and (2)(a). The performance of these heuristics is discussed in Section \ref{sec:results_unproductive}, where this hybrid approach achieves a root-mean-square error (RMSE) of 4.0 cm for linear space prediction made by sweeping. We show this estimate is biased to underestimating space when only perception is used, and that kinesthetic feedback removes this bias in future estimates as it can account for compressibility. 

\subsubsection{Learned Space Estimation}\label{sec:learned-space-estimation}
We have also developed a learning-based model that directly predicts free space from RGB images of the bin and textual metadata of the items within the bin. This model is trained with self-supervised labels derived from the kinesthetic feedback during historical stow events. The architecture employs a Swin-V2 visual encoder \cite{Liu2022CVPR} to process RGB bin images. The text metadata describing items within the bin is tokenized and embedded to match the dimensionality of the visual tokens. These visual and textual representations are fused using a cross-attention mechanism to form a unified representation. A final prediction head maps this representation to a scalar value corresponding to the estimated free-space in physical units. Offline evaluations indicate that this learning-based approach improves predictive accuracy, achieving an RMSE of approximately 2.5 cm in the estimation of free space compared with the 4.0 cm using the heuristic system. Note that this learned space estimation has only been run offline due, and is not included in our deployment results (Section \ref{section:results}).

\section{Behaviors, Motion Planning and Control}\label{section:motion_and_control}

The robotic manipulation system must plan and execute a sequence of complex motions to complete the stowing task. Starting with item grasping at the infeed conveyor, the system proceeds to open the elastic bands on the pod face before generating time-optimal, collision-free trajectories to approach the bin. Then, one of several bin manipulation behaviors is selected and executed to create space and perform controlled placement of the item in-hand. If successful, the robot returns to the infeed conveyor to start the next stow cycle. If stowing was unsuccessful, but the item is still in-hand, a recovery process is executed to recycle the item back into the buffer wall. This multi-stage process requires sophisticated coordination across subsystems and carefully optimized behaviors to maximize throughput and success rate while minimizing amnesty and damage.

\begin{figure}
    \centering
    \includegraphics[width=1.0\linewidth]{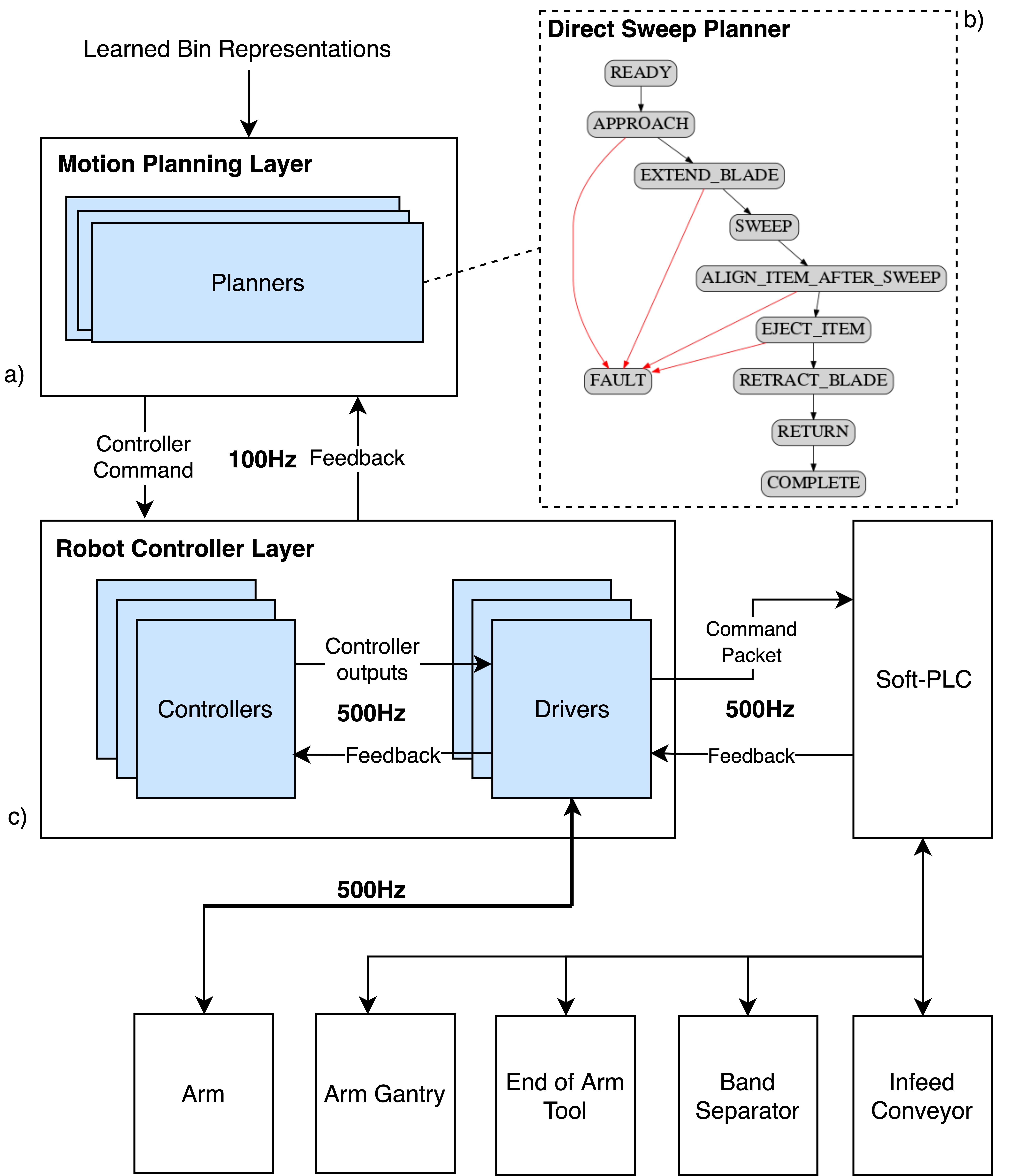}
    \caption{System architecture for motion planning and control: (a) The high-level planning layer uses learned bin representations to generate controller commands. (b) Each task-specific planner is implemented as a state machine composed of reusable \emph{actions}. (c) The lower-level controller layer coordinates multiple hardware subsystems at 500Hz.}
    \label{fig:MotionArchitecture}
\end{figure}

Figure \ref{fig:MotionArchitecture} illustrates the hierarchical motion planning and control architecture. The motion planning layer implements task-specific planners optimized for distinct operations (e.g., elastic band manipulation, sweeping to consolidate in-bin items). Each planner decomposes complex tasks into a sequence of fundamental, re-usable \emph{actions} - atomic motion primitives with configurable parameters that define their execution (e.g., move-to-touch with specified force thresholds, or extend the EoAT plank with defined velocities). These actions generate controller commands to be executed in the controller layer. The commands specify controller type (e.g., joint-space PD, task-space admittance control), gains, and reference trajectories. The motion planner receives controller feedback at 100Hz to determine state transitions and action selection, while the control loop operates at 500Hz to compute and execute appropriate actuator commands based on the current state and desired behavior.

\subsection{Grasp Planning and Execution}\label{section:grasp}
The stowing sequence begins by precisely grasping items from the infeed conveyor. The grasp planner uses item geometry and position data from the infeed perception system to align the EoAT to the item center as shown in Figure \ref{fig:Infeed}. The item is then conveyed into the parallel-jaw gripper, and force control is used to close the gripper and maintain a consistent clamp force through item ejection. Once the item is grasped, the gripper encoders provide an additional measurement of the item's width, improving downstream manipulation planning accuracy. Note that the system currently uses a fixed clamp force of 80N, which can lead to damage on lightweight boxes (see Section \ref{sec:results_damage}).

\subsection{Collision-free Motions Between Infeed and the Bin}
After grasping the item, the manipulator must transport it between the infeed conveyor and target bin while avoiding collisions with the pod and workcell structures. Time optimality of these motions directly impacts the overall stow cycle time. We employ KOMO (k-order Markov Optimization) \cite{toussaint2017tutorial} to generate smooth, collision-free paths, followed by TOPP-RA \cite{pham2018new} optimization to ensure trajectories satisfy joint velocity, acceleration, and torque constraints. To guarantee collision-free motion execution, we implement continuous collision checking between waypoints using sphere-swept convex hull representations of the robot and environment geometries, following an approach similar to \cite{schulman2014motion}.

\begin{table*}[thpb]
    \centering
    \begin{tabularx}{\textwidth}{|>{\raggedright\arraybackslash}X|>{\raggedright\arraybackslash}X|>{\raggedright\arraybackslash}X|>{\raggedright\arraybackslash}X|>{\raggedright\arraybackslash}X|}
    \hline
    \textbf{Direct Insert} & \textbf{Stack} & \textbf{Direct Sweep} & \textbf{Corner Sweep} & \textbf{Item-Push Sweep} \\
    \hline
    \vspace{0.1\baselineskip} No plank insertion or space creation required. Item is inserted vertically into existing free-space. & 
    \vspace{0.1\baselineskip} No plank insertion or space creation required. Item is inserted horizontally into existing free-space. & 
    \vspace{0.1\baselineskip} The plank is inserted vertically between bin wall and adjacent item. Items are consolidated using a hybrid position/force controller while avoiding bin geometry. Item is inserted vertically into the created space. & 
    \vspace{0.1\baselineskip} The plank is rotated and inserted above wall-adjacent item, followed by force-based wall-seeking and wedging to position the plank between wall and item. Remainder of behavior follows Direct Sweep. &
    \vspace{0.1\baselineskip} The plank pushes wall-adjacent item deeper into bin before seeking wall contact in cleared space. After wedging the plank in, the remainder of behavior follows Direct Sweep. \\
    \hline
    \includegraphics[width=3cm]{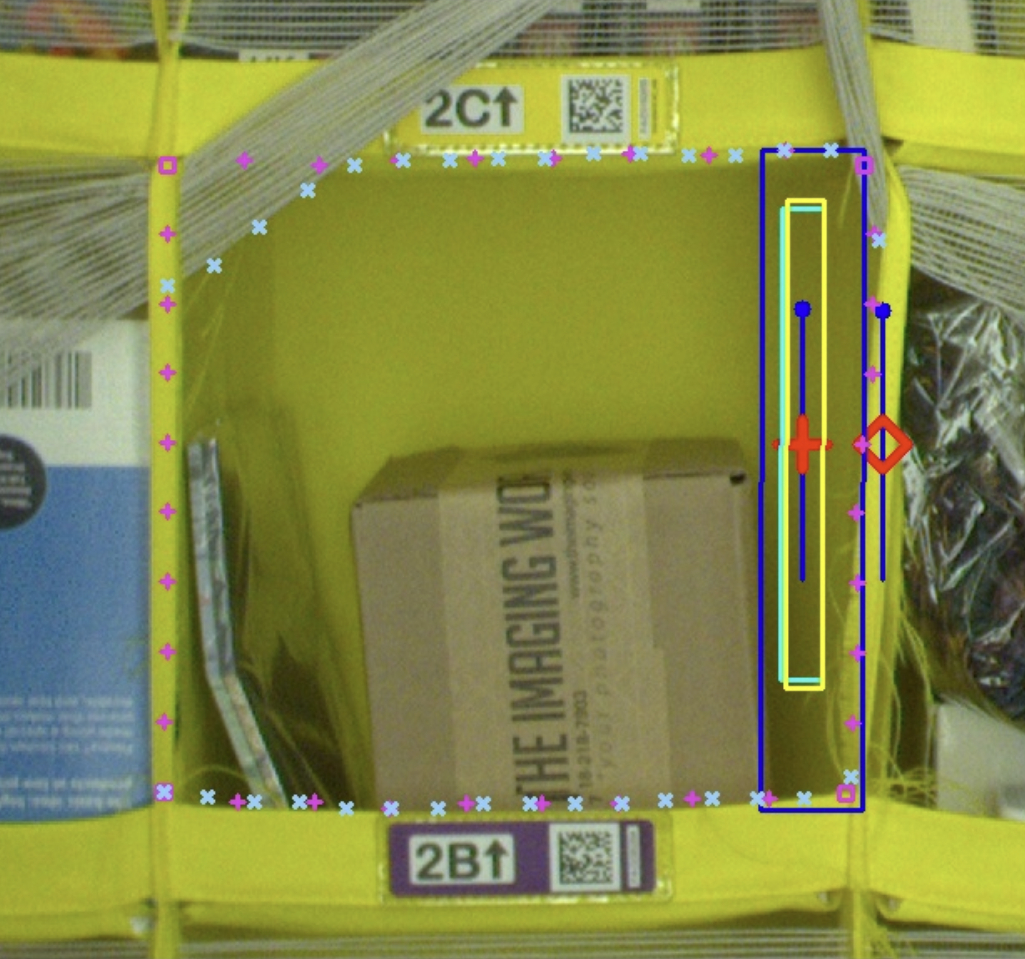} & 
    \includegraphics[width=3cm]{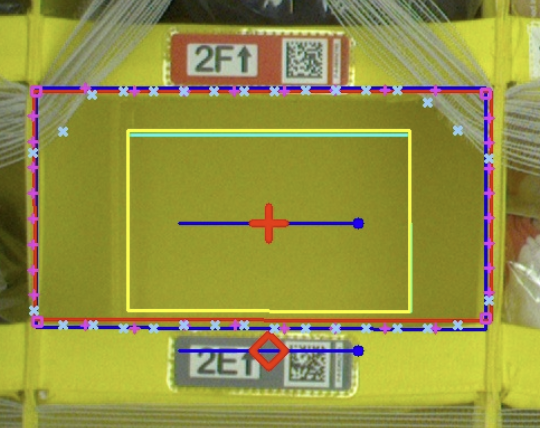} & 
    \includegraphics[width=3cm]{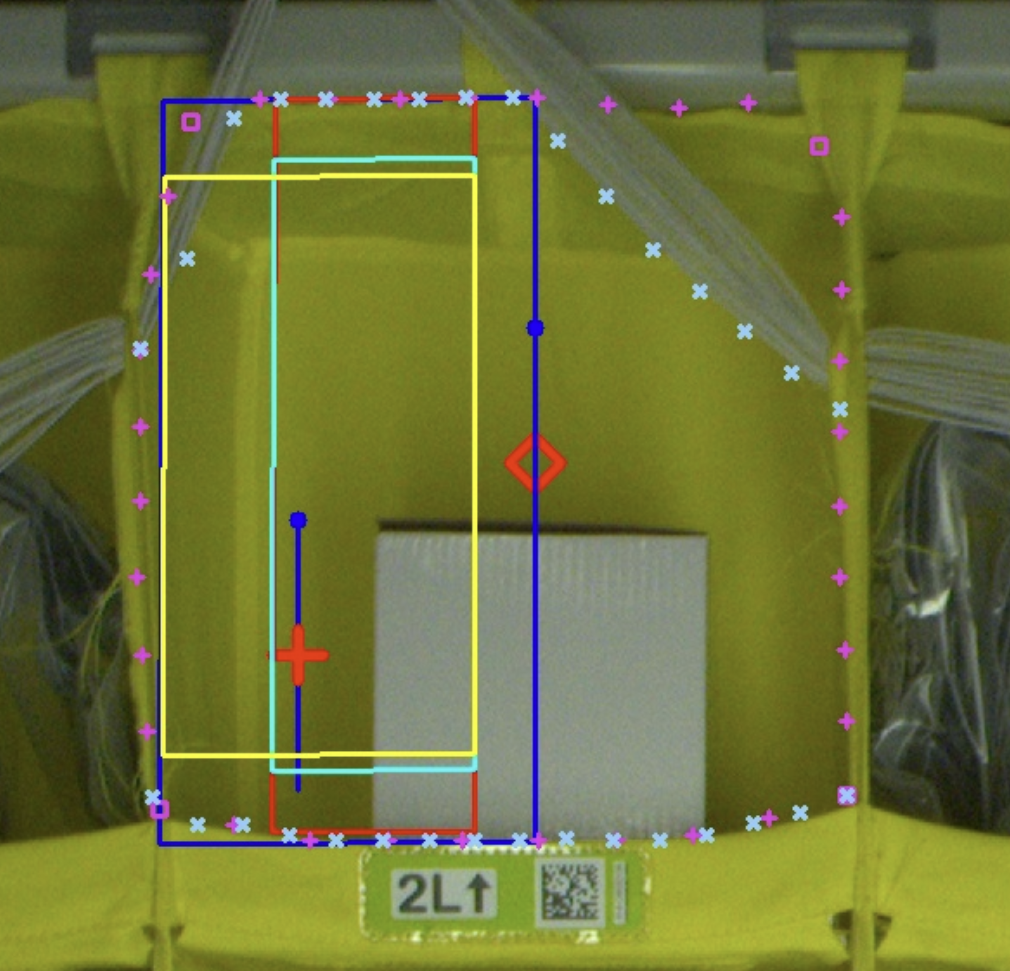} & 
    \includegraphics[width=3cm]{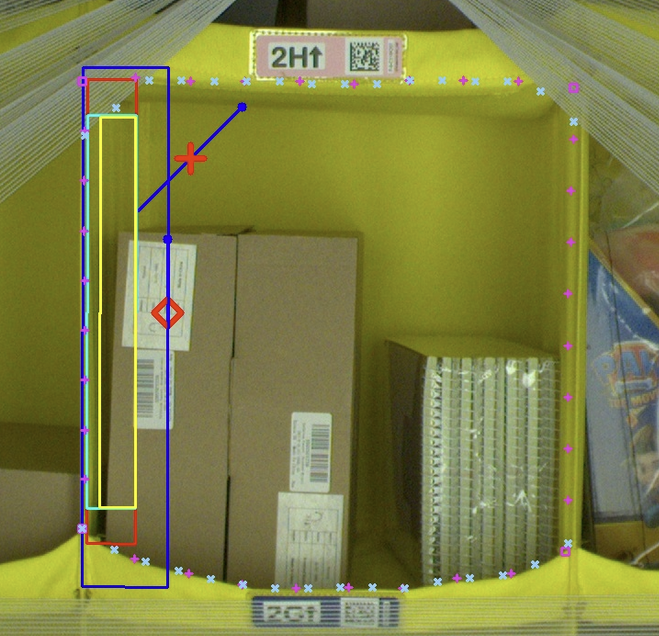} & 
    \includegraphics[width=3cm]{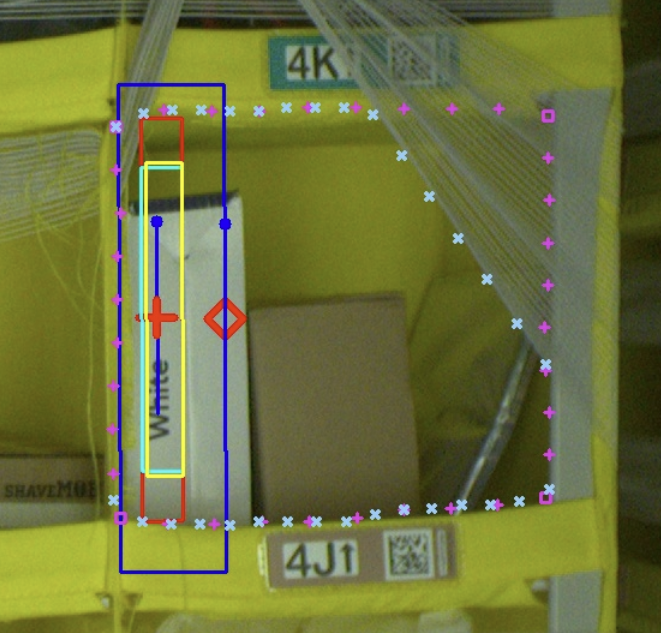} \\
    \hline
    \end{tabularx}
    \caption{Bin manipulation behaviors and example bin states. The blue line represents the plank insertion affordance; the yellow box represents the item insertion affordance.}
    \label{table:behaviors}
\end{table*}

The stowing task benefits from having fixed sets of start configurations (at infeed) and goal configurations (at the pod face). We exploit this structure by pre-computing a lookup table of optimal trajectories through uniform sampling of these configuration sets. Using cloud computing resources, we parallelize the computation to generate approximately 300,000 trajectories in less than 10 minutes. During online execution, the system queries this database and performs local trajectory adjustments to account for slight variations in start/goal poses. This hybrid approach maintains the optimality benefits of KOMO and TOPP-RA while reducing online planning times from seconds to milliseconds.

\subsection{Bin Manipulation Behaviors}\label{sec:behaviors}

\begin{figure*}[t]
    \centering
    \includegraphics[width=\textwidth]{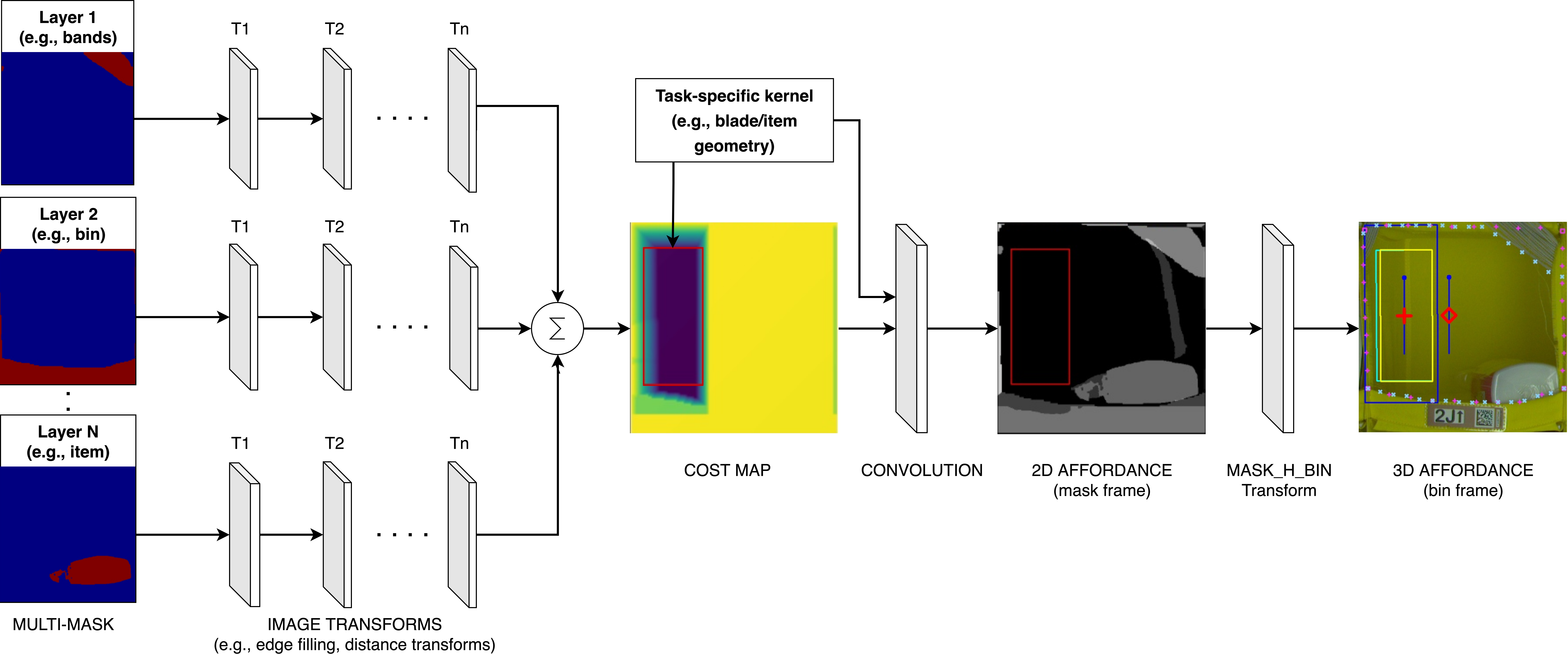}
    \caption{Affordance generation using cost map-based convolution. Individual multi-mask layers are transformed and fused into a 2D cost map. Task-specific kernels representing plank or item geometry are convolved with this cost map to generate optimal 2D pose affordances, which are then transformed to the bin frame for behavior execution.}
    \label{fig:affordance_generation}
\end{figure*}

Once aligned with the target bin, the robotic manipulation system selects from a specialized set of behaviors to organize and pack items efficiently. These behaviors are differentiated primarily by their strategies for three core operations:
\begin{itemize}
    \item \textbf{Plank insertion:} The strategy used to insert the EoAT's thin metal plank into the bin, either directly into free space or by maneuvering around existing items.
    \item \textbf{Space creation:} The process of reorganizing or compressing items within the bin to create room for new items, often involving sweeping motions with the inserted plank.
    \item \textbf{Item insertion:} The final controlled placement of the new item into the created space.
\end{itemize}
Table~\ref{table:behaviors} provides an overview of these key bin manipulation behaviors and example scenarios where each was chosen.

\begin{figure}[tpb]
    \centering
    \includegraphics[width=1.0\linewidth]{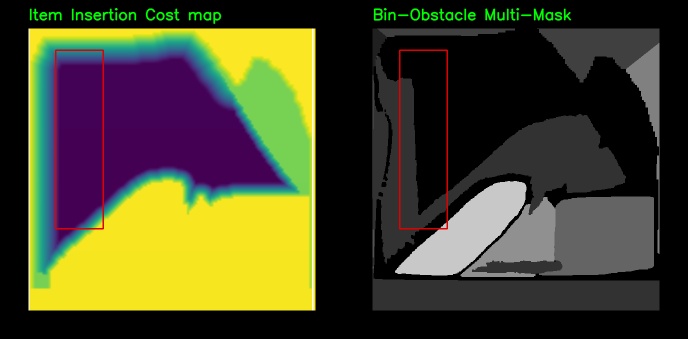}
    \caption{Item-insertion cost map (left) generated from the bin-obstacle multi-mask (right) where the high cost regions are shown in yellow and the low cost regions are shown in blue. The red box indicates the optimal item-insertion pose obtained by convolving the cost map with a kernel with dimensions corresponding to the in-hand item transformed to the mask frame.}
    \label{fig:costmap}
\end{figure}

These behaviors are selected and parameterized by computing \emph{affordances}, which optimize 3D plank or item insertion poses, and consider margins (e.g. between the plank insertion point an object) to ensure that behaviors are executable for a given bin state. Figure~\ref{fig:affordance_generation} outlines the affordance generation framework. First, the layers of the learned bin multi-mask, described in Section~\ref{section:multi-mask}, are transformed using image processing operations (edge filling, distance transforms) to create a 2D cost map that encodes both physical constraints and task objectives. Affordances are generated by convolving task-specific kernels (representing plank or item geometry) with these cost maps. As an example, Figure~\ref{fig:costmap} shows a cost map for the item insertion task where the blue regions indicate more desirable placement locations. The red item bounding box, which serves as the convolution kernel, is shown at the optimal position. When needed, rotated affordances can be generated by simply rotating the kernel before convolution. These affordances are also used in evaluating risk, or the probabilistic outcome of executing a behavior in Section \ref{sec:match}.

To convert 2D affordances into 3D poses, the system projects from the mask frame to the bin frame using it's depth channel. The depth values for bin pixels are extracted and used to reconstruct 3D bin positions in the camera frame. The bin frame-to-mask transform is defined by a 3x4 3D to 2D projection matrix constructed from scaling, rotation, and translation parameters. An inverse 2D to 3D mask-to-bin frame transform is uniquely constructed by using the depth layer of the multi-mask to constrain the missing dimension.

The planners use these plank and item insertion affordances along with robot feedback to construct and update controller commands for behavior execution. During bin manipulation, a hybrid force/position control strategy provides the compliance needed to handle uncertainty while preventing excessive forces that could damage items. For example, in the sweep action, a force-based admittance controller pushes the plank along the primary direction of space-creation (e.g. the y-axis of the pod, defined here as a right to left motion across the bin), consolidating items with a desired force. The remaining degrees of freedom (x, z, roll, pitch, yaw) follow virtual constraints defined by a path indexed to the plank's y-position. This approach enables direct control of sweep forces while maintaining precise plank alignment to avoid collisions with the bin geometry. Kinesthetic feedback during the sweep directly measures available bin space and item compressibility. This information serves two purposes: determining if sufficient space exists for the current item, and updating the perceived free-space model to improve future stow decisions in the same bin.

\subsection{Item Insertion and Exception Handling}\label{sec:insertion-exception-handling}

There are several kinesthetic monitors that run during bin manipulation and item insertion. After the plank has manipulated items in the bin, a kinesthetic space check is run that compares the space created to the width of the item in the gripper. If insufficient space exists, the stow cycle is marked as unproductive and the item is recycled back to the buffer wall via an exception conveyor located below the infeed.

Item insertion is otherwise performed by ejecting the item into the newly created space using the EoAT conveyors. As the item exits the EoAT, the gripper will close under active force control (Section \ref{section:grasp}). 
Conveyor stall detection and gripper position feedback are used to monitor for insertion failures, such as when an item unexpectedly hits the bin wall. Specifically, comparing the gripper position to the expected item width indicates whether the item remains in-hand. If the item is still in-hand, or if any planner has failed before attempting insertion, the stow cycle is also marked unproductive, and the item is retracted back into the hand and is recycled.

These checks allow the system to attempt risky plank insertion strategies and tight item inserts while still being able to retain the item, and return it to the item buffer for another attempt if they fail.

\section{Match Task Planning} \label{sec:match}

The \emph {match} task planning algorithm must decide which items go into which pod bin and what behavior is used to accomplish this task. There are up to 32 items in the buffer wall to choose from, and up to 52 pod bins which can be selected as target destinations. Not all items from the buffer will be stowed into the current pod. While some pods may be relatively empty, others may be almost full, and only one or two items can be stowed into them. The system must also decide when no more items will fit and instead perform a 'kickout', where the mobile robot holding the pod moves away, and another pod is brought in.

In general there is a trade-off between pod density and stow rate, measured in \emph{units-per-hour} (UPH) stored. More items at faster rates can be stowed into empty pod bins compared to cluttered ones. A direct insert behavior does not require any in-bin manipulation and takes on average 10.9 seconds from grasping the item to finishing the cycle. Bin sweeps, where the plank is used to manipulate in-bin items, on average take 13.7 seconds and are also more likely to be unproductive and not stow the item (further timing details and success rates for each behavior are shown in Section \ref{section:results}, Table \ref{tab:primitive_outcome}). An unproductive stow wastes an entire cycle amount of time, and kicking out a pod takes on average 6 seconds before the system can stow again. 

The match planner balances increasing density and stow rate through the notion of \emph{risk}, explicitly modeling the probability that a stow attempt will succeed. Given the generated set of affordances and the features of the items in the buffer (derived from perception and the manifest), match generates a set of feasible \emph{match tuples} $\mathcal{A}$---i.e, elements of  $\mathcal{A}$ are tuples $\{$behavior, item, bin$\}$---where feasibility is determined based on a set of constraints\footnote{The constraints here are related to business logic and other safety considerations related to stowing items, such as not putting heavy items up high in the pod.}. For each stow decision, the match process aims to choose
feasible match tuples that maximally increase expected UPH relative to the current UPH which is computed as a rolling
average. Increasing pod density is implicitly encouraged because minimizing pod kickouts helps maximize UPH, as pod transitions result in unproductive time when no stowing can occur.

Formally, let $(X,a)\in \mathcal{X}\times \mathcal{A}$ denote the current state and a given action $a\in \mathcal{A}$ (e.g., the selection of a bin-item-primitive match tuple), and let $Y$ be the random variable representing a stow outcome where $Y=1$ indicates 'success' and $Y=0$ indicates 'failure'. For state-action pair $(X,a)$, the expected UPH for outcome $Y$  is given by
\begin{multline}\label{eq:euph}
        \mathbb{E}[\UPH(Y)|X,a]=\Pr(Y=1| X,a)\frac{N_s+1}{T+{t}(Y=1| X,a)}\\
    + \Pr(Y=0| X,a)\frac{N_s}{T+{t}(Y=0| X,a)},
\end{multline}
where  $\Pr(Y| X,a)$---i.e., the \emph{risk model}---is the probability of outcome $Y$, the cycle time of outcome $Y$ is $t(Y| X,a)$, the cumulative stow time for the given period is $T$,  and the number of successful stows for the given period is $N_s$.  
Observe that $\Pr(Y| X,a)$ and $t(Y| X,a)$ are unknown \emph{a priori}, and hence we must estimate them using learned models. 

\subsection{Frequentist Risk Model}\label{sec:frequentist_risk}
A simple frequentist risk model was developed wherein we rank by the empirical success of each behavior, while adhering to strict geometric constraints. In particular, the system prunes out stows such that the item dimensions are within a heuristically set margin of the available space.

The system then starts with tuples containing the highest ranked behavior and re-ranks that subset with a ``largest item in smallest available space" heuristic, comparing the item-bin pair in each tuple. It then greedily sequences all these tuples, before repeating this process for lower ranked (i.e. higher risk) behaviors. This allows the system to maximize the space utilization within each pod while choosing the match tuples whose primitive has the highest probability of success. 

Practically this means that direct insert behaviors will be prioritized when there is sufficient margin (as they are the most successful on average), and the largest item heuristic means we stow as close to that margin as possible. After all possible direct inserts are planned, bin sweeps and other behaviors are used to exploit the additional space they can create. 

\subsection{Learned Risk Model}\label{sec:learning_risk}
Learned risk models that consider the behavior, item and bin state tuple explicitly can provide an improvement over prioritizing behaviors by type as per the previous approach. These models can be learned from experience, predicting the probability of success and the cycle time, which are then used in \eqref{eq:euph} to predict the expected UPH. 

These models also use exploration as part of their training regiment. An epsilon-greedy \cite{Sutton1998RL} like approach is used. While most training data was captured under the frequentist match algorithm in the prior section, up to 5\% of the time, the system relaxes margins and/or randomly chooses a behavior for which there is an affordance to explicitly provide training data beyond the low-risk, high-reward action space. 

The learned risk model is a tabular gradient boosted decision tree classifier that is trained and calibrated on engineered features; the score is then utilized to predict the probability of success and failure. The tabular data includes features that encode the geometric fit (e.g., item height to bin height margin) as well as item and primitive attributes and the current bin state (e.g., detected available linear width) obtained from perception.  For example, Figure~\ref{fig:costmap} shows the item insertion cost map where the item bounding box (red) is overlaid on the multi-mask which abstracts the bin state (i.e., locations of obstacles in the bin). Darker colors indicate available space; the engineering features encode a measure of `fit' of the item (bounding box) within the available space. For cycle time, we forecast the failed and successful cycle times separately; each such predictor is a tabular decision tree regressor based on bin and primitive engineered features. 

We show in Section \ref{sec:results_uph} that this learned model provides ~7\% boost in stow rate over the frequentist model, but it does come with a cost of needing to continually recollect training data as development evolves and behaviors change.

\section{Deployment Results} \label{section:results}

\begin{figure*}[htbp]
    \centering
\begin{overpic}[width=\textwidth]{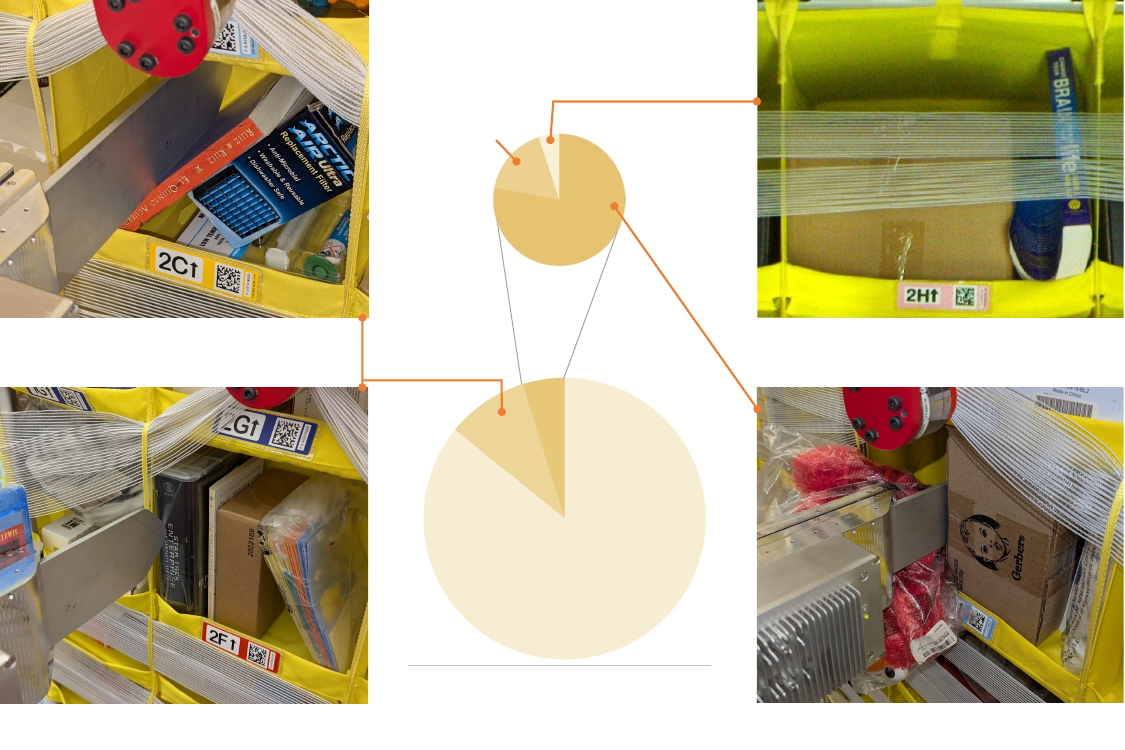}
    \put(3,35){\parbox{5cm}{\centering \small{a) The items jammed during a bin sweep which failed to make space.}}}
    \put(3,1){\parbox{6cm}{\centering \small{c) Plank insert failed to push a long item back and contact with the bin wall.}}} 
    \put(70,35){\parbox{5cm}{\centering \small{b) A book page bent during eject.}}} 
    \put(70,1){\parbox{5cm}{\centering \small{d) An object deformed during ejection and fell to the ground. }}}
    
    \put(44,15){\parbox{5cm}{Successful (86\%)}}
    \put(53,59){\parbox{5cm}{Damage (0.2\%)}}
    \put(34,35){\parbox{2cm}{\centering Unproductive (9.3\%)}}
    \put(55,47){\parbox{2cm}{\centering Amnesty (3.7\%)}}
    \put(37,55){\parbox{2cm}{Other (0.8\%)}}
\end{overpic}
\caption{Stow outcomes over 100,000 attempts.}
\label{fig:outcomes}
\end{figure*}
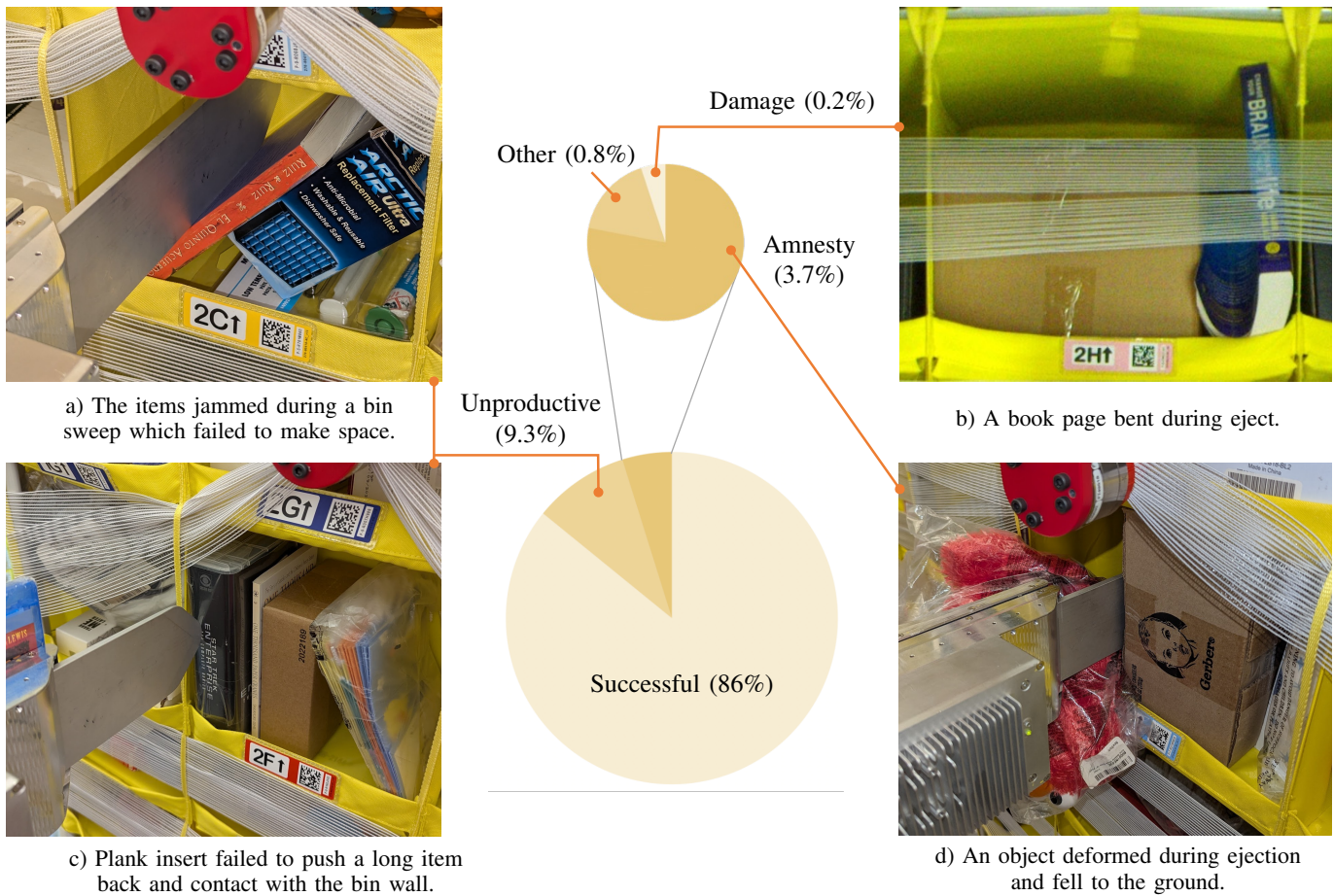

The robotic stow system was deployed in an large e-commerce warehouse, and outcomes for the most recent 100,000 stow attempts are analysed. All stow outcomes were manually annotated by human observers and are classified as one of four outcomes: (1) A stow \emph{success} is when the target item is stowed correctly and there is no damage or amnesty. (2) An \emph{unproductive} stow cycle is when the item is not successfully inserted, but it remains securely in the EoAT and the item can be recycled to the item buffer for future stow attempts. No damage or amnesty occurs during an unproductive stow. (3) \emph{Amnesty} is defined as either the grasped item or an item already in the pod falling to the floor. (4) \emph{Damage} includes stows where the system induced any defects on the grasped item or items already in the pod. Note that unproductive and amnesty outcomes are automatically labeled through item tracking, but all outcomes reported here were validated by human annotators. Examples of these outcomes and their frequencies are shown in Figure \ref{fig:outcomes}. For the remainder of this section we use \emph{defects} to refer to any amnesty, damage or off-nominal outcome noted by annotators that should be remediated. This last category, labeled as \emph{other} in Figure \ref{fig:outcomes}, is a subjective catchall for many events, including items overhanging out of the bin or bands not closing properly, which have the potential to cause issues later as the pods move around.

Table \ref{tab:primitive_outcome} shows these stow outcomes further broken down by bin manipulation behavior, as described in Section \ref{sec:behaviors}. The ``direct insert" and ``stack" behaviors achieve high success rates as they only attempt stows into free space, and do not require manipulating objects in the bin. Failures in these behaviors primarily stem from calibration errors and uncertainty in the grasped item pose and shape (e.g., for deformable items) during tight insertions. The sweep behaviors show lower success rates due to the challenges of contact-rich manipulation.

\begin{table}[htbp]
\centering
\caption{Stow outcomes over 100,000 attempts.}
\begin{tabular}{|l|l|r|r|l|}
\hline
\textbf{Bin Manipulation} & \textbf{Outcome} & \textbf{Count} & \textbf{Percent} & \textbf{Avg. Cycle} \\
\textbf{Behavior} & & & & \textbf{Time (s)} \\
\hline
\multirow{3}{*}{\shortstack[l]{All}} 
 & Successful & 85,859 & 85.86\% & --  \\
 & Unproductive & 9,311 & 9.31\% & -- \\
 & Defect & 4,830 & 4.83\% & -- \\
 & \ -- \ Amnesty & 3,767 & 3.77\% & -- \\
 & \ -- \ Other & 819 & 0.82\% & -- \\
 & \ -- \ Damage & 244 & 0.24\% & -- \\
\hline
\multirow{3}{*}{\shortstack[l]{Direct Inserts \& \\Stack}} & Successful & 72390 & 90.72\% & 10.87 \\
 & Unproductive & 4530 & 5.68\% & 18.24 \\
 & Defect & 2875 & 3.60\% & 10.17 \\
\hline
\multirow{3}{*}{Bin Sweeps} & Successful & 13469 & 66.67\% & 13.68 \\
 & Unproductive & 4780 & 23.66\% & 19.69 \\
 & Defect & 1955 & 9.68\% & 16.83 \\
\hline
\end{tabular}
\label{tab:primitive_outcome}
\end{table}

\subsection{Unproductive Stows}\label{sec:results_unproductive}

Plank insertion failures are a leading cause of unproductive sweep cycles. As shown in Figure \ref{fig:outcomes}b, items positioned against the bin wall pose particular challenges. The ``item push" behavior attempts to create clearance by pushing in-bin items deeper before seeking wall contact for plank insertion. However, rigid, long items that cannot be displaced lead to insufficient clearance and missed wall contacts. Additionally, deformable items can conform around the plank, making wall detection difficult. 

Another major source of unproductive sweep cycles is when the behavior fails to create the amount of space predicted by perception. These failures can occur due to occlusion of items, heavy items requiring excessive force to move, or deformations in the bin wall geometry from opening the bands. Figure \ref{fig:outcomes}a shows an example where a blister pack was partially hidden below the bin lip and the clear plastic of the blister pack contained undulations which caused the other items to jam during compression with the plank. 

Bin sweeps gain kinesthetic information during exection, that is used to re-estimate the space in the bin as described Section \ref{sec:space_estimation}). This kinesthetically informed space estimate was used in selecting  35\% of sweeps overall and provides improved space estimates. Perception--only space estimates under predict space by 36mm on average with a standard deviation of 40mm. Kinesthetically informed estimates were less biased, over predicting space by 0.15mm on average with a standard deviation of 34mm.  

The match algorithm also attempts riskier stows to optimize rate, trading off the risk of failure with stowing more items to the pod and avoiding pod kickout. Unproductive stows are thus expected even in more mature versions of this robotic system, but will aim for much lower rates (i.e. 3\% vs. 9\%). Exploring, which executes riskier stows, will continue to be critical for continual learning and maintaining robust models.

\subsection{Stow Rate} \label{sec:results_uph}

The stow robot \emph{rate} is comparable to that of a human. Over the month of March 2025, humans stowed at an average rate of 243 units per hour (UPH) while the robotic systems stowed at 224 UPH. This comparison was careful to compare human stowers operating on the same floor as the robotic workcells, as stow rates vary based on inbound item distributions and the density of items already in the fabric pods. It was also found that humans had greater variation in stow rates: people can quickly stow many small items efficiently, but are slower with large items, crouching for lower bins, or when using a step ladder to reach the top bins. It is estimated that using the robot stow system to populate only top rows of pods would increase human stow rates by 4.5\% overall and would avoid the use of step ladders.

The above UPH results were achieved using the frequentist model match algorithm (Section \ref{sec:frequentist_risk}). A/B testing, outside of the 100,000 stow attempts, was used to show that stow rates are improved by approximately 7\% using the learned risk models from Section \ref{sec:learning_risk}. Here the frequentist model is used as the 'control', and the learned risk algorithm is the 'treatment', where the randomization unit is at the pod level---i.e., we randomize the treatment over the pods upon arrival. We hypothesize that the treatment outperforms (with statistical significance) the control---i.e., the null hypothesis is $H_0:\ \{\mu_{A}\geq \mu_{B}\}$, 
where $A:= {\tt Alg}_{\mathrm{freq}}$ and ${B}:={\tt Alg}_{\mathrm{learned}}$, and $\mu_{\tt T}$ (respectively, $\mu_{\tt C}$) is the mean UPH for the treatment (respectively, control) over the duration of the experiment.  

\begin{table}[h!]
    \centering
    \begin{tabular}{|l|l|c|c|c| }
    \hline
         Test&Algorithm& $n$ (\# pods)& Avg ${\UPH}$ &Avg ${\UPH}\pm 95$\%CI\\\hline\hline
     A &${\tt Alg}_{\mathrm{freq}}$ & 695 & 307.6 & (313,302)\\ \hline
     B &  ${\tt Alg}_{\mathrm{learned}}$ & 227 & 326 & (336,316)\\ \hline
    \end{tabular}
    \caption{A/B test where we randomized the treatments over the pods that arrive to a single workcell. With significance level $\alpha=0.01$, we reject the null hypothesis $H_0=\{\mu_{A}\geq \mu_{B}\}$ with $p$-value $0.008$.}
    \label{tab:ab_test_1}
\end{table}
The frequentist model uses fixed risk and heuristic decision making.
On the other hand, the learned risk naturally adapts the risk tolerance based on system performance: if the number of successful stows $N_s$ is low due to limited space in the arriving pods, the system will naturally take on more risk. This framework also allows for optimally choosing between slow but high probability of success behaviors vs fast but low probability of success behaviors.

\subsection{Amnesty}

Failure to separate the bands completely is the leading cause of amnesty. This occurred on 4.8\% of all stows, either because the band separator simply misses the bands, or more commonly, because the bands are pinned down at the edges and overlap over with the item insertion area. While this impacts all metrics, it caused 19\% of amnesty cases, where clipping the band can sometimes 'sling-shot' the item back out of the bin after it is let go. The system currently continues to stow if band overlap is detected after opening, as 60\% of stows in this condition still succeed. From a rate perspective, most of the cycle time is in grasping and getting the item to the bin, so the marginal time to attempt the insert is low (1-4 seconds). This behavior needs to be changed as the system scales, and an abort model that determines the outcome of item-band interactions is being developed.

In general, amnesty has similar root causes to that of unproductive stows, but the system does not detect that there is a lack of space or that the item has not been inserted properly (see Section \ref{sec:insertion-exception-handling}), and continues to stow. This often occurs for thin or deformable items where it is difficult to interpret kinesthetic information. Figure \ref{fig:outcomes}d shows a soft toy in a plastic bag that was dropped after it was ejected. The toy caught on the bin as it was ejected, deformed downwards, and slid along the bin edge rather than going into the bin. The compressible nature of the item made it difficult to sense this kinesthetically. Similarly, unsuccessful item release is also difficult to detect when stowing thin items ($\lesssim $ 2mm) due to compliance in the EoAT's parallel jaws.

\subsection{Damage}\label{sec:results_damage}

Damage is a high cost outcome for the system that can only occur at very low levels before the system is unviable commercially. The relatively low damage rate ($\sim$ 0.2\% of stows) can largely be attributed to avoiding item-to-item contact during insertion, as the EoAT design separates in bin items from the inserted item with the extendable plank. Damage tends to occur when this ideal is violated, and the item instead is pushed into other items, bands or the walls of the bin. The EoAT pinch gripper action also avoids box deconstruction common in suction grippers \cite{Aduh2024ICRA}, and the conveyance mechanism has produced negligible damage rates. 

The top cause of damage ($>$25\%) in the system is from amnesty, where items impacting the floor causes damage. The next leading causes of damage is specific to books, where they are damaged during insertion (14\%), as shown in Figure \ref{fig:outcomes}b. Other leading causes of damage included crushing lightweight boxes in the EoAT (10\%) and damaging items in the bin with the plank (6\%).

While some sources of damage as easy to mitigate, others are more complex. For example, the crushing of items can likely be mitigated by adjusting grip force based on item weight. Most items that were crushed were very lightweight, and lower grip force would still retain the item. On the other hand, Figure \ref{fig:outcomes}b shows damage that was caused during the insertion of a book -- the pages of the book hit the adjacent object and are folded backwards. This complex contact interaction is both item specific and can occur with very small overlaps. Given the prevalence of books in this e-commerce warehouse, creating a special case set of margins or behaviors may be warranted. More generally margins and grasp force could be set using item context, for instance using LLMs as in \cite{Xie2024Deligrasp}.

\subsection{Observations and Outlook}

\emph{Robot morphology makes this problem tractable:} A human stower uses complex bimanual hand motions, typically moving the band and in-bin items dexterously with one hand and using their other hand to stow the object. Humans also contact and slide the in-hand object against other objects in the bin, much like you would in placing books on a densely packed shelf. The stow robot decomposes the problem differently using dedicated hardware to solve each sub-task (see Section \ref{section:hardware}): a band separator to manipulate the bands; an extendable EoAT plank to manipulate in-bin items; and conveyance within the EoAT paddles to insert the item. This decomposition means the inserted item is not being used to create space, which removes item-to-item dynamic interactions and largely shields the item from damage (see Section \ref{sec:results_damage}). An independent band separator allows item grasping to occur in parallel, meaning that while a human stower is still faster at in-bin manipulation, stow rates are comparable between the two (see Section \ref{sec:results_uph}). 

\emph{Engineered or heuristic systems can be used for bootstrapping learned models:} The results presented in this paper use learned perception to infer a bin state. Engineered behaviors and task planning are used to sequence and execute motions. These heuristic methods achieve reasonable performance levels (i.e. total system success $>$85\%) and can also generate data for learned systems. It is believed that further improvements will require deeper use of object properties in both moderating behaviors and predicting  outcomes. 

Learning behaviors and task planning is a logical next step. In the context of this paper, learned space estimation in Section \ref{sec:learned-space-estimation} and learned risk prediction in Section \ref{sec:learning_risk} were compared to the baseline system and found to improve upon it.  We note that compared to learning pick points from simulation scene data \cite{Newbury2023}, stowing requires understanding item compressability or deformability, and both the in-hand and in-bin items can deform and move in unexpected ways. This motivated the use of using real robotic data. Furthermore, there is a larger action space and when training models, we often found we had collected data for only a small manifold of the total space. For instance, the baseline system puts the largest item in the smallest space with a margin constraint, and the vast majority of samples were thus close to this margin. Extending beyond this meant including deliberate exploration and development of exception handling strategies (section \ref{sec:insertion-exception-handling}.

\emph{Both kinesthetic and visual feedback are required for stowing diverse items:} The stow robot currently uses visual perception for computing affordances (e.g. what behaviors can be executed), predicting task success and task planning, whereas kinesthetic feedback is primarily used in task execution. Kinesthetic behaviors allows for fast operation, but visual cues are likely needed during behavior execution to remove the long tail of amnesty and insertion failure cases. In particular, deformable objects do not consistently transmit force, making it difficult to monitor task progress using kinesthetic information only. For instance, in Figure \ref{fig:outcomes}d, the toy could be re-detected or tracked during insertion to understand that it was bending or deforming outside of the target insertion point. This could be used to abort the insert, allow models to predict that state more directly, or be used in feedback policies.

\emph{Not all failures are the same:} This robotic system is now being deployed at larger scales in an e-commerce warehouse. While the system has demonstrated human like stow rates and can maintain the flow of items into the storage floor, an increased focus on reducing defects is still required. Unproductive cycles, where the robot fails to stow the item, only cost time, whereas amnesty or damage required human remediation. Further scaling will require a disproportionate focus on reducing defects. 

\section{Conclusion}
This paper presents an end-to-end robotic system designed for autonomous stowing of items into fabric pods in production warehouse environments. The system was deployed in an e-commerce fulfillment center and has conducted over 500,000 stows at greater than 85\% success at a similar stow rate to humans. Notably, the system uses visual perception to estimate space in cluttered, occluded bins. It uses the concept of 'risk' to choose bins to interact with and to optimize stow rates. For cluttered bins, the system uses an extendable plank on the end effector and a set of behaviors to sweep items to one side, making space. Kinesthetic feedback is then used to refine space estimates, and can be used for offline training of models. 

Deploying this system and analyzing its failure modes gives insights into what can be expected operating with large and diverse item sets. There are object classes (like books) that can be problematic with induced damage. Deeper research into predicting complex item interaction may be required in scaling manipulation more generally. To aid in that endeavor, a test data set of bin images, item manifests, resulting stow outcomes, and kinesthetically sensed space creation outcomes has been published and shared with the community\footnote{We are planning to release the dataset to the public upon approval after the paper is in review}.

\bibliographystyle{IEEEtran}
\bibliography{stow}

\end{document}